\documentclass[sigconf]{acmart}

\pdfoutput=1
\usepackage{graphicx}
\usepackage{amssymb}
\usepackage{amsmath,amsfonts}
\usepackage{algorithmic}
\usepackage{graphicx}
\usepackage{textcomp}
\usepackage{xcolor}
\usepackage[ruled,linesnumbered]{algorithm2e}
\usepackage{threeparttable}
\usepackage{multirow}
\usepackage{tabularx}

\newcommand{\nop}[1]{}

\usepackage{subcaption}
\usepackage[ruled]{algorithm2e}

\AtBeginDocument{%
  }

\setcopyright{acmlicensed}
\copyrightyear{2018}
\acmYear{2018}
\acmDOI{XXXXXXX.XXXXXXX}

\acmConference[Conference acronym 'XX]{Make sure to enter the correct
  conference title from your rights confirmation emai}{June 03--05,
  2018}{Woodstock, NY}
\acmISBN{978-1-4503-XXXX-X/18/06}

\renewcommand\footnotetextcopyrightpermission[1]{}
\begin{document}

\title{Up-sampling-only and Adaptive Mesh-based GNN for Simulating Physical Systems}



\author{Fu Lin}
\orcid{0009-0001-9342-8414}
\email{2231531@tongji.edu.cn}
\affiliation{%
  \institution{Tongji University}
  \city{Shanghai}
  \country{China}
}

\author{Jiasheng Shi}
\email{shijiasheng@tongji.edu.cn}
\affiliation{%
  \institution{Tongji University}
  \city{Shanghai}
  \country{China}
}

\author{Shijie Luo}
\email{sjlaw@tongji.edu.cn}
\affiliation{%
  \institution{Tongji University}
  \city{Shanghai}
  \country{China}
}

\author{Qinpei	Zhao}
\email{qinpeizhao@tongji.edu.cn}
\affiliation{%
  \institution{Tongji University}
  \city{Shanghai}
  \country{China}
}

\author{Weixiong Rao}
\email{wxrao@tongji.edu.cn}
\affiliation{%
  \institution{Tongji University}
  \city{Shanghai}
  \country{China}
}

\author{Lei Chen}
\email{leichen@cse.ust.hk}
\affiliation{%
  \institution{Hong Kong University of Science and Technology (GZ)}
  \city{Guang Zhou}
  \country{China}
}


\begin{abstract}
Traditional simulation of complex mechanical systems relies on numerical solvers of Partial Differential Equations (PDEs), e.g., using the Finite Element Method (FEM). The FEM solvers frequently suffer from intensive computation cost and high running time. Recent graph neural network (GNN)-based simulation models can improve running time meanwhile with acceptable accuracy. Unfortunately, they are hard to tailor GNNs for complex mechanical systems, including such disadvantages as ineffective  representation and inefficient message propagation (MP). To tackle these issues, in this paper, with the proposed Up-sampling-only and Adaptive MP techniques, we develop a novel hierarchical Mesh Graph Network, namely UA-MGN, for efficient and effective mechanical simulation. Evaluation on two synthetic and one real datasets demonstrates the superiority of the UA-MGN. For example, on the Beam dataset, compared to the state-of-the-art MS-MGN, UA-MGN leads to 40.99\% lower errors but using only 43.48\% fewer network parameters and 4.49\% fewer floating point operations (FLOPs). 
\end{abstract}

\maketitle

\section{Introduction}
\begin{figure}[t]
  \centering
  \includegraphics[width=.9\linewidth]{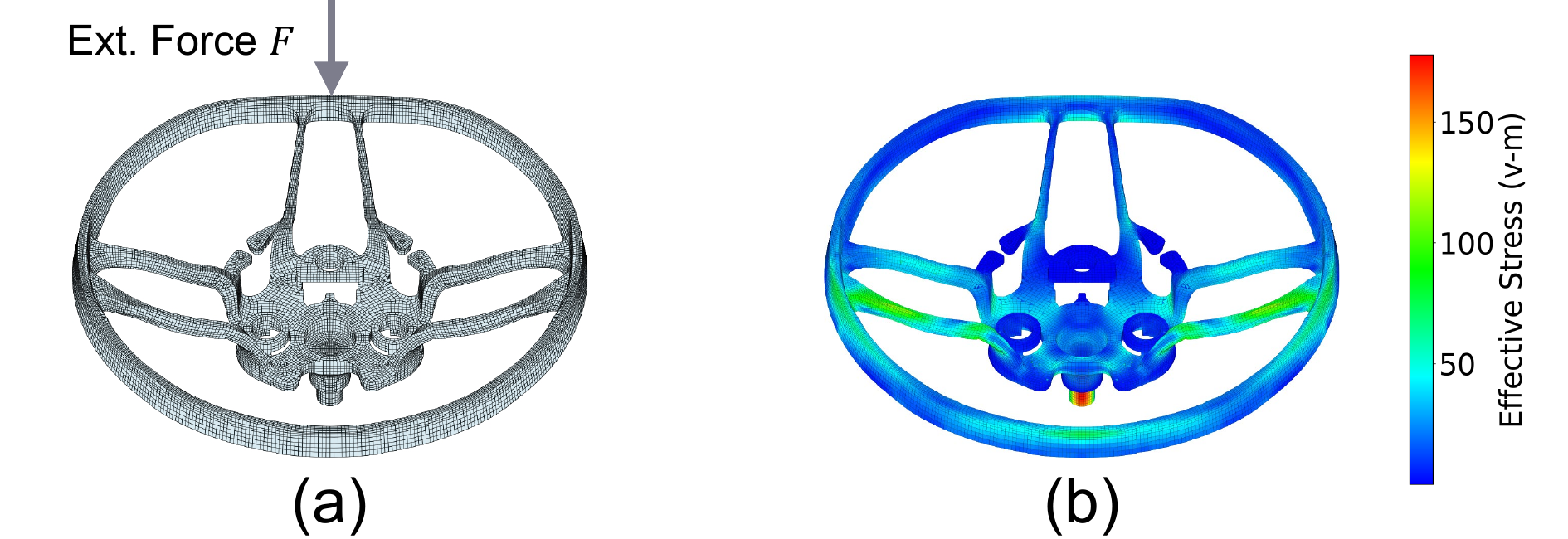}
  \caption{Simulation of a steering wheel fixed on a bottom plane. (a) Initial input: geometric structure and external force $F$. (b) Resulting stress field computed by a FEM solver.}\label{fig1}   \vspace{-2ex}
\end{figure}

Simulation of complex mechanical systems is important in numerous engineering domains, such as structural mechanics \cite{ NGUYENVAN2023103333,FU2023} and aerodynamics \cite{KOU2021100725, BAZILEVS202324, STRUCHKOV20207}. Traditional simulations rely on the numerical solution of Partial Differential Equations (PDEs), e.g., using the Finite Element Method (FEM). For example, given an external force $F$, Figure \ref{fig1}
illustrates the FEM simulation result of stress field on a steering wheel when the steering column is fixed on the bottom plane. To perform mechanical simulation, FEM tools first divide the input steering wheel into a mesh structure (e.g., a triangular mesh of roughly equal triangle size) and derive the {numerical} solution of stress field. When the number of divided mesh elements is high (e.g., tens of thousands and even more), we have to solve a large number of PDEs, suffering from intensive computation cost and high running time. Moreover, when simulation boundary conditions (e.g., the force $F$ or the geometric structure of the steering wheel) change, the FEM solver has to re-process the entire simulation, leading to high overhead.

Recently, with the success of deep learning, researchers have developed end-to-end learning models that map simulation input to the output results \cite{KOCHKOV2021e2101784118,GREENFELD2019,RAISSI2019686}. The adopted models include Convolutional Neural Networks (CNNs) \cite{NILS202025,DESHPANDE2022115307,ZHENGUO2019011002} and Graph Neural Networks (GNNs) \cite{YASMIN2024,ALLEN2022,LAM20231416}. Compared to FEM numerical solvers, learning models improve running time meanwhile with acceptable simulation accuracy. In particular, when the input {geometric} object is divided into a mesh structure (such as a triangular mesh), the works \cite{PFAFF2020,FORTUNATO2022} model the mesh structure as a graph and exploit GNNs to learn a spectrum of mechanical system simulation.

Unfortunately, existing works are hard to tailor GNNs for effective and efficient simulation of complex mechanical systems. {Above all}, effective representation of a mechanical system, i.e., the steering wheel above, is non-trivial, involving the overall global representation and accurate local one, e.g., the small area to which the external force $F$ is applied. (1) Existing works perform node feature representation equally for all nodes in the mesh graph, with \emph{no differentiation of those boundary nodes} where the external force $F$ is applied. If the number of boundary nodes is rather small among all mesh nodes (see Figure \ref{fig1}), no differentiation of boundary nodes could  falsely miss the associated meaningful node features and suffer from inaccurate local representation. (2) {Typical Internet and social networks exhibit power law graphs, and the GNNs on these graphs frequently require 2 or 3 message propagation (MP) steps \cite{LI2023126441,9046288,10.1145/3580305.3599562}}. Instead, the \emph{node degree distribution of mesh graphs is rather even}, ranging from 3 to 40 with an average 5 - 6 in our datasets. How to tune the MP steps in mesh graphs is difficult. Too high MP steps lead to the over-smoothing issue \cite{QIMAI2018,CHEN20203438}, making the representation of mesh graph nodes rather similar and indistinguishable. Too small steps may not propagate the effect of the external force $F$ from a small number of its boundary nodes to the entire steering wheel. (3)  MP on too many small triangles in a triangular mesh could lead to \emph{message propagation loops} surrounding such triangles, again leading to the over-smoothing issue \cite{Rong2019DropEdgeTD,10195874}. 

Besides the ineffectiveness issue above, existing GNN works 
on complex mechanical system simulation suffer from \emph{high computation overhead}. More specifically, for better global and local representation of a mechanical system (i.e., the steering wheel), multi-level mesh graphs have been developed \cite{FORTUNATO2022,LINO2021,ZONGYI2020, PERERA2024117152}, where the bottom graph is with the finest mesh (e.g., the smallest triangle size) and the top one is the coarsest mesh (with the largest triangle size). Hierarchical GNN models, e.g., the popular U-shaped networks \cite{FORTUNATO2022,LINO2021,ZONGYI2020}, can effectively learn the multi-level mesh graphs. However, these hierarchical GNN models require the MP {to start} from the finest mesh graph (the bottom one) to coarse ones until the top one, and next back to the original bottom one, involves both down- and up- sampling steps. Obviously, the two sampling steps require high MP overhead and suffer from the inefficiency issue.

To tackle the aforementioned issues, in this paper, we develop a novel Up-sampling-only and Adaptive Mesh Graph Network ($\mathsf{UA}$-$\mathsf{MGN}$), consisting of two key techniques. Firstly, we propose an up-sampling-only GNN model on multi-level mesh graphs. That is, it performs MP firstly on the top coarse mesh graph and next fine ones with gradually increasing resolutions until the bottom finest one. It means that our UA-MGN model requires the up-sampling steps alone. Unlike the previous works requiring both up- and down- sampling steps, our model does not require the down-sampling and thus leads to {a} much smaller number of MP steps. Moreover, since it performs MP firstly on coarse mesh graphs, those boundary nodes have {a} chance to be within GNN receptive fields at the \emph{early} stage for better global representation. Meanwhile, it postpones the propagation of too many local {messages} on fine mesh graphs at the late stages, and thus our UA-MGN model can avoid the inefficient propagation of those local {messages} on fine mesh graphs and mitigate the over-smoothing issue.

Secondly, we develop an adaptive MP tailored to diverse mechanical systems. That is, in a mesh graph, we first divide mesh edges into groups based on edge directions (e.g., by the  $K$-means algorithm), next tune the MP steps for every edge group, and then perform the adaptive MP. 
After that, we perform adaptive propagation along the grouped edge directions by the associated {MP} steps. Intuitively, it indicates some edge directions may involve much further MP and others are limited to small areas. Thus, this technique not only mitigates the issues of infinite MP loops and over-smoothing, but also is adaptive to the diverse mechanical simulation with various geometric shapes and external conditions for better generalization. As a summary, we make the following {contributions} in this paper.

\begin{itemize}
    \item To the best of our knowledge, this is the first to tailor a hierarchical learning model for efficient and effective simulation with up-sampling-only GNNs and adaptive MP.
   \item We develop the up-sampling-only hierarchical GNN model, leading to higher efficiency and better global receptive fields {at} the early stage.
    \item Our adaptive technique
    can tune the number of MP steps  along grouped edge directions to overcome the issue of infinite MP loops and over-smoothing, and leads to high simulation generalization capability.
    \item Evaluation on two synthetic and one real datasets demonstrates the superiority of our work UA-MGN. For example, on the Beam dataset, compared to {state-of-the-art MS-MGN \cite{FORTUNATO2022}, UA-MGN leads to 40.99\% lower errors meanwhile with higher efficiency by using 43.48\% fewer network parameters and 4.49\% fewer floating point operations (FLOPs)}.
    
\end{itemize}
The rest of this paper is organized as follows. Firstly, Section \ref{sec2:relate} reviews related works and Section \ref{sec:problem} next gives the problem definition. Then, Section \ref{sec4:solution} presents solution details. After that, Section \ref{sec5:experiment} evaluates our work, and Section \ref{sec6:conclude} finally concludes the paper.

\section{Related Work}\label{sec2:relate}

\textbf{CNN-based Simulation}. CNN-based approaches have been recognized for their effectiveness in many simulation works, due to the capacity to learn spatial and temporal dependencies from data. For example, some previous works demonstrate the application of CNNs in fluid simulation \cite{XIAOXIAO2016481,NILS202025}, and some exhibit their power in material mechanical simulation \cite{ZHENGUO2019011002,IBRAGIMOVA2022103374}. Nonetheless, CNNs suffer from issues {in representing complex mechanical systems and irregular geometric structures.

\textbf{GNN-based Simulation}. GNNs have emerged as a promising solution to model the topologies and interactions of mechanical systems. The work \cite{ALVARO2020} demonstrates that GNNs can effectively learn dynamic interactions in particle-based fluid systems by representing 
{neighbor} particles as {connected} graph nodes. The previous work \cite{PFAFF2020} exploits GNNs to simulate various mechanical systems by leveraging mesh graphs to represent the geometric structure of such systems. Yet, flat GNNs in these works do not work well {in representing} complex geometric structures due to the limited range of MP. To address this issue, some works \cite{FORTUNATO2022,PERERA2024117152} introduce hierarchical GNNs to extend the range of MP via low-resolution meshes with a larger mesh size. The recent work \cite{PERERA2024117152} employs skip connections by MP on a multi-graph consisting of uniform grids with various resolutions, and information can be spread beyond the local neighborhoods. However, these approaches require careful trade-off between the range of MP and representation resolution.

\textbf{Neural Operators for PDE Solutions}. Instead of end-to-end neural networks, some works \cite{LI2020NO,ZONGYI2020,PATEL2021113500} propose to replace {compute-intensive} PDE operators by neural networks. In this way, neural operators are incorporated into the PDE computation framework. However, the performance of neural operators still depends upon the original PDE operation, typically leading to much higher overhead, when compared to end-to-end CNN or GNN-based models. The Fourier neural operator \cite{LI2020} mitigates this issue by employing frequency domain multiplications via Fourier transforms, as an alternative to spatial domain integrals. Such an operator can optimize global representation, but at the cost of worse local precision and spatial interactions. Fourier transforms perform well only on uniform grids. To overcome this limitation, the very recent work \cite{10.5555/3648699.3649087} introduces learnable deformation from physical irregular meshes to computational uniform grids in the geometric domain. However, this work still faces challenges of complex topologies when there does not exist a diffeomorphism from the physical space to the computational space \cite{10.5555/3648699.3649087}.


\textbf{Physics-Informed Models}. Unlike the approaches above, some works \cite{RAISSI2019686,10.1145/3580305.3599835,MENG2020113250} introduce physical information into solution frameworks by incorporating PDEs into loss functions, known as Physics-Informed Neural Networks (PINNs). Moreover, the works \cite{10.1145/3580305.3599466,SUN2020112732,10.1007/s00466-020-01928-9} incorporate PDEs into neural networks by carefully designing with expert knowledge. 
PINNs can reduce the demand for the amount of training data and enhance physical interpretability. 
Nevertheless, when the PDEs change (e.g., the geometric domain changes), these PINN methods have to re-adjust the loss function or neural networks, and re-train the entire models.

\section{Problem Definition}\label{sec:problem}
\textbf{PDEs for mechanical systems}. 
In literature, PDEs are ubiquitous in mathematically oriented scientific fields, such as engineering and physics, and have been widely used to model mechanical systems, involving initial and boundary conditions. Here, the \emph{initial conditions} indicate the state of the system at the initial time step $t=0$, and the \emph{boundary conditions} define the behavior of the system at the boundary of the geometric domain, such as the geometric structure of the simulation object and {the} external force. The PDEs and the associated initial-boundary conditions define the dynamic evolution of system states. By the previous work  \cite{Epstein2017}, we define the following PDE governing a mechanical system.
\begin{equation}\small
    \mathbf{F}\left ( \mathbf{x}, t; u, 
    \frac{\partial u}{\mathbf{x}}, 
    \frac{\partial^{2} u}{\partial \mathbf{x}^{2}}, ..., 
    \frac{\partial u}{t}, 
    \frac{\partial^{2} u}{\partial t^{2}}, ...   \right )=0, \quad
    \mathbf{x} \in D,
    0\le t \le T
\end{equation}
where $u\left ( \mathbf{x}, t \right )$ denotes the function to be solved involving the space coordinates $\mathbf{x}$ within the geometric domain $D$ and time $t$, and $\mathbf{F}$ is a {function regarding a certain mechanical system}. If $\mathbf{F}$ is a linear function of $u$ and its derivatives, the PDE is said to be linear. Regarding the initial-boundary conditions, we give the following example.

\begin{equation}\small
\begin{matrix}
u\left ( \mathbf{x}, 0 \right )=u_{0}\left ( \mathbf{x}\right ), \quad \mathbf{x} \in D,\\
u\left ( \mathbf{a}, t \right )=f_{a}\left ( t\right ),
u\left ( \mathbf{b}, t \right )=f_{b}\left ( t\right ), \quad \mathbf{a},\mathbf{b} \in bd
\end{matrix}
\end{equation}

Here, $u_{0}\left ( \mathbf{x}\right )$ is the initial state for $\mathbf{x} \in D$ with $t=0$, and for two certain point sets $\mathbf{a}$ and $\mathbf{b}$ at the {boundary {of $D$} denoted by $bd$}, the two functions $f_{a}\left ( t\right ), f_{b}\left ( t\right )$ indicate the time-dependent boundary behaviors of the two boundary point sets $\mathbf{a}$ and $\mathbf{b}$, respectively.
For the example in Figure 1, the set $\mathbf{a}$ could denote the mesh nodes to which the external force $F$ is applied, and $\mathbf{b}$ {could denote} the nodes {at the bottom of the steering column} which is fixed on the plane.


\textbf{Finite Element Method (FEM)}. FEM stands as a cornerstone of numerical methods to solve PDEs, and has been widely used as de facto ground truth in many mechanical engineering applications \cite{2013i}. Practically, FEM first discretizes an input simulation object with continuous shapes or bodies in the geometric domain $D$ to a set of divided mesh elements. Depending on a specific application, the mesh elements could be either surface elements (i.e., triangles) or volume elements (i.e., tetrahedra). After the discretization, we have a mesh graph $G=\{V, E, C\}$, where $V=\left \{ \mathbf{v}_{i}  \right \}, \mathbf{v}_{i} \in D$ is the set of mesh nodes, $E=\{\{\mathbf{v}_{i}, \mathbf{v}_{j}\}\}$ with $i \ne j $ is the set of mesh edges, 
and $C=\left \{c_i \right \}$ is the set of mesh elements which are surrounded by mesh nodes and edges. Each element $c_i$ is a subdomain of $D$ (e.g., triangles in a triangular mesh, or tetrahedra in a tetrahedral mesh) and $\bigcup_{i}{c_{i}} = D$.
Given the discretized mesh elements, we can exploit FEM to solve the PDEs and have numerical results of $u\left ( \mathbf{v}_{i}, t \right )$.

\begin{definition}\label{problem1}
[Mesh Graph-based Mechanical Simulation] Given a mechanical system modeled by a mesh graph $G=(V,E, C)$ with an initial condition $u_{0}\left ( \mathbf{x}\right )$, boundary conditions $f_{a,b,...} (t )$ and the resulting response $U=\{u\left ( \mathbf{v}_{i}, t \right )\}$ with $\mathbf{v}_{i} \in V$, we want to learn a mesh graph regression model $R(\cdot)$ with $U= R(G,u_{0}, f_{a,b,...})$.
\end{definition}
In the problem above, the simulation input includes the mesh graph $G$ and initial-boundary conditions {applied to} a subset of nodes $V$, for example, an external force $F = \{ \mathbf{f}_k\}, v_k \in V$ at $t=0$. The simulation output is a set of mechanical responses $u\left ( \mathbf{v}_{i}, t \right )$, indicating the state of each node $\mathbf{v}_i$ at time step $t$. Depending upon the specific simulation application, we might be interested in the detailed simulation result of continuous time steps (e.g., fluid simulation) or the final convergent simulation result for a {large} time step $t$ (e.g., rigid body simulation).

Note that the problem definition above requires that simulation systems can be modeled as mesh graphs. For those fluid and rigid body simulations, we can comfortably exploit nowadays FEM tools to generate mesh elements and next model them as a mesh graph. Yet for those simulation systems such as human mobility, traffic control and urban city behaviour system simulations \cite{ZhangYJL22,abs-2405-12520,10.1145/3534678.3539440}, it is non-trivial to model such systems within the geometric domain $D$ by mesh graphs and we may resort to other techniques, i.e., multi-agent and discrete event simulation.

\begin{figure}
  \centering  \includegraphics[width=.9\linewidth]{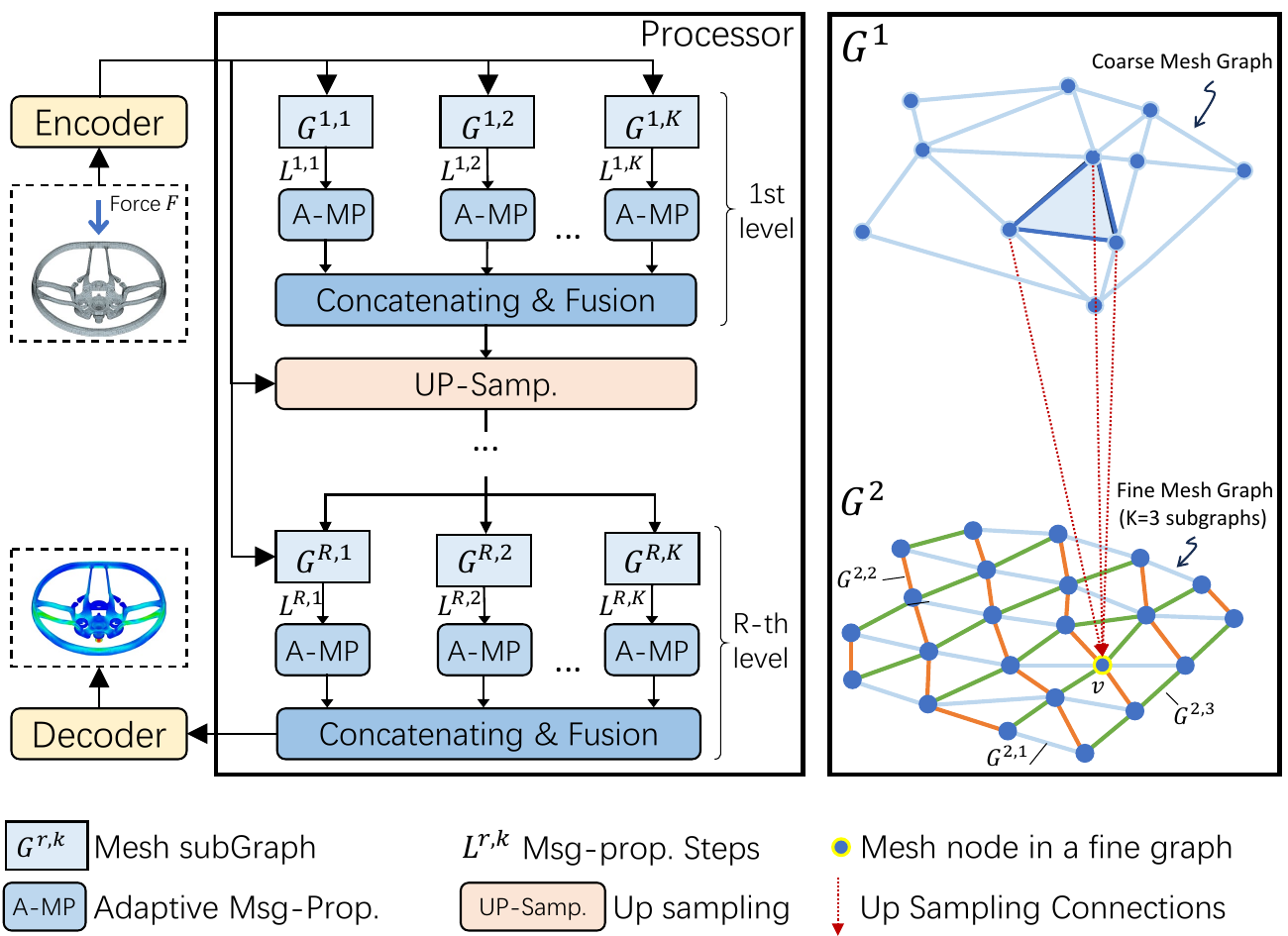} \vspace{-2ex}
  \caption{Overview of UA-MGN framework}\label{fig2}  \vspace{-4ex}
\end{figure}

\section{Solution Detail}\label{sec4:solution}
\subsection{Overview}
Before introducing our solution detail, we first give $R$-level mesh graphs that are required by UA-MGN. In Figure \ref{fig2}, the bottom $R$-th mesh graph is with the {smallest} mesh element size and the top one is the coarsest mesh with the largest element size. Note that the bottom mesh graph $G^{R}$ is just the input graph $G$ given by Problem \ref{problem1}, i.e., $G^{R} = G$. To generate  coarser mesh graphs, say $G^{r}$ with $1\leq r \leq R-1$, we could iteratively set a larger mesh element size than the one of the current graph $G^{r+1}$ to perform a lower resolution discretization on the geometric domain $D$, e.g., {by the widely used Delaunay triangulation \cite{10400424}}. By repeating the mesh generation step, we can have the top coarsest mesh graph $G^{1}$. Practically, we set $R=3$ that is {sufficient} to learn global and local representation, and a larger $R$ may lead to a very large mesh element size in the top mesh graph and suffer from the {mesh fragment} issue \cite{WORDENWEBER1984285}. Now given the $R$-level mesh graphs $G^{1}, ..., G^{R}$, we can build an associated GNN for each mesh graph, and expect that the MP on the top coarsest graph $G^{1}$ can learn the global representation of the mechanical system and yet the MP on the bottom finest graph $G^{R}$ captures the local representation of a small area in the system. With {the} help of the $R$-level mesh graphs, we have {a} chance to capture both global and local representation of the system.

\subsection{Up-sampling-only Graph Neural Networks}
{Overall}, our UA-MGN network follows the Encoder-Processor-Decoder framework.
The key is that the Processor stage requires the up-sampling-only steps from the top coarsest graph to the bottom finest one across the multi-level mesh graphs.

\textbf{Encoder}. The encoder learns each of $R$-level mesh graphs, including mesh edges and nodes, into embedding vectors. Since mesh nodes and edges are with spatial coordinates, we expect that the learned embedding vector should be independent {of} specific coordinate systems that are adopted by FEM solvers, i.e., the so-called \emph{shift invariance} or \emph{spatial invariance} \cite{BRONSTEIN2021,HAN2020}. We thus exploit relative position coordinates of edge endpoints, instead of absolute ones. To learn {edge embedding vectors}, for an undirected edge $\{\mathbf{v}_{i}, \mathbf{v}_{j}\}$, we regard it as two directed edges: $\vec{\mathbf{e}}_{ij}$ from $\mathbf{v}_{i}$ to $ \mathbf{v}_{j}$ and $\vec{\mathbf{e}}_{ji}$ from $\mathbf{v}_{j}$ to $ \mathbf{v}_{i}$, and develop the edge feature for each directed edge. {By taking $\vec{\mathbf{e}}_{ij}$ for illustration, we have one relative displacement vector $\mathbf{v}_{j}-\mathbf{v}_{i}$, and the norm $|\mathbf{v}_{i}-\mathbf{v}_{j} |$, i.e., the Euclidean distance of $\vec{\mathbf{e}}_{ij}$. 

For each node $\mathbf{v}_{i}$, we mainly focus on those nodes at the boundary of the geometric domain or within the initial-boundary conditions. That is, we have two node features: a binary indicator $\mathbf{I}_{i}$ equal to 1 if  $\mathbf{v}_{i}$ is at the boundary and otherwise 0, and values regarding the initial-boundary conditions. For example, in the input graph $G$, if 
an external force $F = 100$ Newtons is applied to an area of 5 mesh nodes, we assume that each node is on average with 20 Newtons. Next, since such node features are originally provided by the finest input graph $G^R=G$ but not by the coarse graphs $G^1, ..., G^{R-1}$, we {thus} interpolate these node attributes from the input graph $G^R$ to the coarse graphs $G^{1}, ..., G^{R-1}$, e.g., by {the  barycentric interpolation \cite{doi:10.1137/S0036144502417715}}. Now, for each mesh graph, given the mesh edge features and node features, we exploit a MultiLayer Perceptron (MLP) to transform the concatenated features into a latent vector of size 128.

\textbf{Processor}. This step is to update the node and edge embedding vectors by (1) adaptive MP among the nodes within each mesh graph $G^r$ for $1\leq r\leq R$, and (2) up-sampling operations from coarse graphs $G^r$ to fine ones $G^{r+1}$ until $G^R$.

For the MP within the graph $G^r$, 
we perform \emph{adaptive message propagation} along graph edges. That is, depending upon the  average number of edges of the mesh element, we choose the number $K$ and divide $G^r$ into $K$ subgraphs $G^{r,1}, ..., G^{r,K}$ (identified by $K=3$ colors in Figure \ref{fig2}), and perform the proposed adaptive MP (that will be given in Section \ref{sec4.2}). That is for each subgraph $G^{r,k}$ with $1\leq k\leq K$, we have the associated MP steps  $L^{r,k}$. In this way, we do not propagate messages from a certain node evenly to all neighbours, and instead perform directed MP adaptively along graph edges. For each {MP} step, we perform the embedding update:
\begin{equation}\vspace{-1ex}\small
\vec{\mathbf{e}}^{r}_{ij}\gets f_E^{r} ( \vec{\mathbf{e}}^{r}_{ij}, \mathbf{v}^{r}_{i}, \mathbf{v}^{r}_{j} ), \quad\quad
\mathbf{v}^{r}_{i}\gets f_V^{r} (\mathbf{v}^{r}_{i}, \sum_{j}\vec{\mathbf{e}}^{r}_{ij}  ) 
\end{equation}

Next, the key of our multi-level UA-MGN model is to build the \emph{up-sampling connections} from the nodes in coarse mesh graphs $G^r$ to those in fine graphs $G^{r+1}$. Thus, for every node $\mathbf{v}_j$ in fine graphs $G^{r+1}$, we first need to locate  a coarse mesh element $\mathbf{c}^{r}_{i}\in G^{r}$, where the node $\mathbf{v}_j$ belongs to, and next build connections from every vertex node of the located coarse mesh element $\mathbf{c}^{r}_{i}$ to {the node} $\mathbf{v}_j$. In the right subfigure in Figure \ref{fig2}, for the node, say $v$, in $G^2$, we can first locate a triangle element in the coarse graph $G^1$ which this node $v$ belongs to, and next build three up-sampling connections from  three nodes of the found triangle element to the node $v$. Such a projection ensures that every node in a fine mesh graph can receive the up-sampling operation from the nodes in a coarse graph.

Denote $\vec{\mathbf{e}}^{r,r+1}_{ij}$ to be an up-sampling connection from a vertex node $\mathbf{v}_i \in G^{r}$ in the found mesh element $\mathbf{c}^{r}_{i}$ to a node $\mathbf{v}_j \in G^{r+1}$. We perform the following up-sampling MP operation to update a node embedding vector $\mathbf{v}^{r+1}_{j}$ in $G^{r+1}$:


\vspace{-2ex}
\begin{equation}\small
\begin{matrix}
\vec{\mathbf{e}}^{r,r+1}_{ij}\gets f^{r,r+1}_{E} (\vec{\mathbf{e}}^{r,r+1}_{ij},\mathbf{v}^{r}_{i},\mathbf{v}^{r+1}_{j}  ), \quad
\mathbf{v}^{r+1}_{j}\gets f^{r,r+1}_{V} (\mathbf{v}^{r+1}_{j},\sum_{i}\vec{\mathbf{e}}^{r,r+1}_{ij})
\end{matrix}
\label{equation4}
\end{equation}

We again implement $f^{r,r+1}_{E}$ and $f^{r,r+1}_{V}$, $r=1, \dots, R-1$ by MLPs {with an output embedding size of 128}.

\textbf{Decoder}. This step again exploits the MLP to transform the node embedding vectors  $\mathbf{v}^{R}_{i}$ in the $R$-th level finest mesh graph back to the output mechanical response (such as stress field on the entire mesh graph $G$).

\subsection{Adaptive Message Propagation}\label{sec4.2}
For a given mesh graph $G^r$, the adaptive message propagation (MP) within $G^r$ involves two tasks. Firstly, we need to divide $G^r$ into $K$ subgraphs $G^{r,k}$ with $1\leq k\leq K$, and next tune the MP steps  $L^{r,k}$ on the subgraph $G^{r,k}$. Since each subgraph $G^{r,k}$ is with an associated number $L^{r,k}$, we expect to guide the MP purposely towards those important subgraphs but not equally towards all subgraphs.

\textbf{Mesh Graph Division}. To enable the directed MP along mesh edges, our general idea is to cluster those edges with similar directions into the same group. That is, if the included angle of two edges is close to zero, i.e., two edges are parallel, we would like to cluster them into the same group. Following the idea, we exploit the classic $K$-means algorithm to divide the edges of a certain mesh graph into $K$ subgraphs. Intuitively, we now have $K$ different edge directions associated with such divided subgraphs.

\begin{algorithm}[htbp]
    \caption{Divide mesh graph $G$ into $K$ subgraphs}
    \label{algorithm1}
    \KwIn{Mesh graph $G=(V,E,C)$ and the number $K$}
    \KwOut{$K$ subgraphs $G^{1}, \ldots, G^{K}$}
    
    Init. $K$ groups $E^1, \ldots, E^K$ with random directions $\vec{\mu}^1, \ldots, \vec{\mu}^K$\;
    
    \While {the clustering {stopping} condition is {not satisfied}} {
        \ForEach{edge $\mathbf{e}_{ij} \in E$} {

            Set $\vec{\mathbf{e}}_{ij} = \mathbf{v}_{j} - \mathbf{v}_{i}$ as the direction vector of edge $\mathbf{e}_{ij}$\;
           
            Compute $\theta^k = \mathrm{arccos} ( \frac{\vec{\mathbf{e}}_{ij} \cdot \vec{\mu}^k} {\|\vec{\mathbf{e}}_{ij}\| \|\vec{\mu}^k\| }   )$ for $k=1,\ldots,K$\;
            
            Assign edge $\mathbf{e}_{ij}$ to cluster $E^k$ with the min. $\theta^{k}$\;
        }
        
         \lForEach{cluster $1\leq k\leq K$}{update $\vec{\mu}^k$ by mean($E^k$)}
    }
    \Return{$G^{1}=(V, E^1), \ldots, G^{K}=(V, E^K)$}
\end{algorithm}

Alg. \ref{algorithm1} gives the Pseudocode of mesh graph division. Line 5 computes the distance of an edge $\vec{\mathbf{e}}_{ij}$ and a cluster $\vec{\mu}^k$ by the arc-cosine function, and line 8 computes the mean of a cluster by the average of all edge direction vectors within the cluster. Finally, when the cluster membership does not change, we stop the loop in line 2.

{Time complexity} is $O(T\cdot K\cdot |E|)$ where $T$ (resp. $K$) is the number of iterations (resp. groups) and $|E|$ is the edge count.

\textbf{Tuning MP steps}. For each divided subgraph $G^{r,k}$, in Alg. \ref{algorithm2}, we tune an associated number of MP steps $L^{r,k}$ by using the largest diameter of a certain connected component (CC) within the subgraph $G^{r,k}$. To this end, we first project every coarse element $\mathbf{c}^{r-1}\in G^{r-1}$ into the fine graph $G^{r}$ (see Figure \ref{fig3}). That is, for every node $\mathbf{v}_j$ in $G^{r}$, we can locate a coarse element $\mathbf{c}^{r-1}$ to which $\mathbf{v}_j$ belongs, e.g., by BVHTree \cite{Collision}.
The nodes {belonging} to the same coarse element and their neighboring edges then form an area $\mathbf{a}^{r}$ in $G^{r}$, such that the area $\mathbf{a}^{r}$ can cover the coarse element $\mathbf{c}^{r-1}$.

\begin{figure}[h]
  \centering  \includegraphics[width=.82\linewidth]{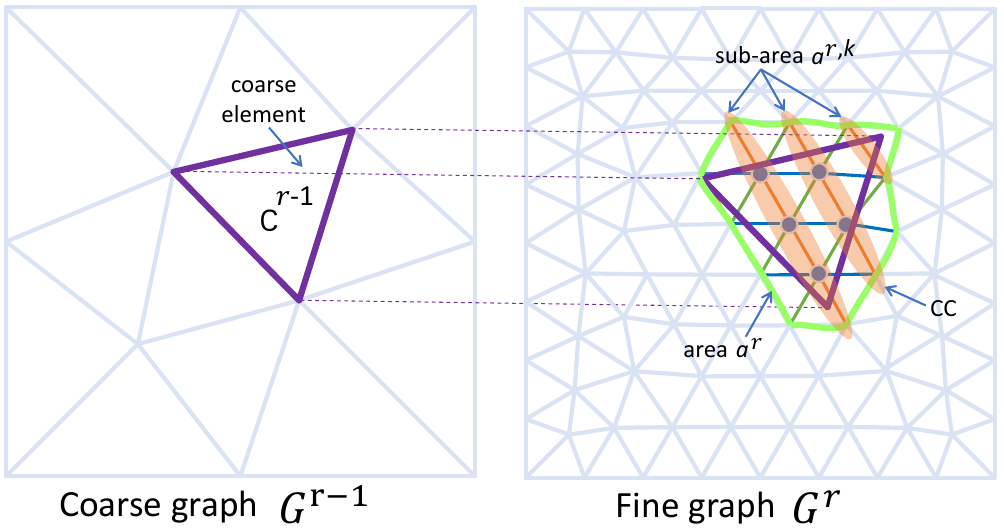}\vspace{-2ex}
  \caption{Tuning the Number of MP steps  $L^{r,k}$}\label{fig3}  \vspace{-2ex}
\end{figure}

Note that we have already divided the mesh graph $G^{r}$ into $K$ subgraphs $G^{r,k}$ with $1\leq k\leq K$. As a result, the area $\mathbf{a}^{r}$ is further divided into multiple sub-areas $\mathbf{a}^{r,k}$. In lines 6-9, for each sub-area $\mathbf{a}^{r,k}$, we may have multiple {CCs} and next find the diameter $l^{r,k}$ for each {CC}. For a certain $1\leq k\leq K$, a coarse element $\mathbf{c}^{r-1}\in G^{r-1}$ is with the largest diameter $l^{r,k}$ on the fine graph $G^{r}$, and we thus can find the {largest} one $L^{r,k}$ among all coarse elements in $G^{r-1}$. In Figure \ref{fig3}, the area $a^r$ is with three sub-areas (due to $K=3$ groups), and the sub-area $a^{r,k}$ highlighted by the orange color is with three CCs. Among such CCs, we can find that the largest diameter is 4. 

It is not hard to find that the found number $L^{r,k}$ is the diameter regarding a certain {CC} within some {sub-areas} $\mathbf{a}^{r,k}$. Since the size of such a {CC} is smaller than the size of the sub-area $\mathbf{a}^{r,k}$, much smaller than the size of the area $\mathbf{a}^{r}$, and significantly smaller than the subgraph $G^{r,k}$, we thus have {a} chance to greatly optimize the overhead of MP by setting a small number $L^{r,k}$ of {MP} steps. Alg. \ref{algorithm2} lists the steps to tune the number $L^{r,k}$.

Time complexity: The running time mainly depends upon the projection (line 3) with $O(|V| log|C|)$ and the computation of diameters (line 7) with $O(|V|+|E|)$, leading to the total 
$O(|V| log|C| + |E|)$, where $|C|$ is the number of coarse mesh elements in $G^{r-1}$.

\begin{algorithm}[htbp]
    \caption{Tune the MP steps  $L^{r,k}$}
    \label{algorithm2}
    \KwIn{ $G^{r-1}, G^r$ and $K$ subgraphs $G^{r,1}, \ldots, G^{r,K}$ }
    \KwOut{ $K$ numbers of {MP} steps $L^{r,1}, \ldots, L^{r,K}$}
    Initialize $L^{r,k} \gets 0$ for $1\leq k\leq K$;
    
    \ForEach{element $\mathbf{c}_{r-1} \in G^{r-1}$} {
        Find the area $\mathbf{a}^{r}$ by projecting $\mathbf{c}_{r-1}$ to $G^r$;
        
        Divide $\mathbf{a}^{r}$ into $K$ sub-areas $\mathbf{a}^{r,k}$ by $G^{r,1}, \ldots, G^{r,K}$;

        \ForEach{sub-area $\mathbf{a}^{r,k}$} {
            \ForEach{Connected Component $cc$ in $\mathbf{a}^{r,k}$} {
            Compute the diameter $l^{r,k}$ of $cc$;

                \lIf{$l^{r,k} > L^{r,k}$} {
                    Update $L^{r,k} \leftarrow l^{r,k}$
                }
            }
        }
    }
    \Return{$L^{r,1}, \ldots, L^{r,K}$}
\end{algorithm}

\textbf{Adaptive Message Propagation}. Until now, we are ready to give the adaptive MP within each subgraph $G^{r,k}$ by the steps $L^{r,k}$ to update node and edge embedding vectors.
\begin{equation}\small
\begin{matrix}
 \vec{\mathbf{e}}^{r,k,l+1}_{ij}\gets f_E^{r,k} ( \vec{\mathbf{e}}^{r,k,l}_{ij}, \mathbf{v}^{r,k,l}_{i}, \mathbf{v}^{r,k,l}_{j} ), \\
\mathbf{v}^{r,k,l+1}_{i}\gets f_V^{r,k} (\mathbf{v}^{r,k,l}_{i}, \sum_{j}\vec{\mathbf{e}}^{r,k,l+1}_{ij}  ), \\
k=1\dots K; \quad l=0\dots L^{r,k}-1
\end{matrix}
\end{equation}
\begin{equation}\small
\mathbf{v}^{r}_{i}\gets g_V^{r} ( [ \mathbf{v}^{r,1,L^{r,k}}_{i}, ..., \mathbf{v}^{r,K,L^{r,k}}_{i}  ] 
 )
\label{equation6}
\end{equation}

In the equations above, we first update the edge and node embedding vectors by the $l$-th {step of} MP within each subgraph $G^{r,k}$ with totally $L^{r,k}$ steps (see Eq. 5), and next concatenate and aggregate all node embedding vectors of $K$ subgraphs (see Eq. 6).
{Here, we implement $f_E^{r,k}$, $f_V^{r,k}$, $g_V^{r}$ by MLPs {with an output embedding size of 128}, and the network parameters of $f_E^{r,k}$ and $f_V^{r,k}$ are shared within the same {subgraph}.}
The aggregated vectors are then fed into the mesh graph $G^r$ for $1\leq r\leq R-1$ via the up-sampling step, and finally to the decoder in the $R$-th level finest mesh graph $G^R$. During these steps, we can find that each node embedding is updated by (1) the MP within the subgraph $G^{r,k}$ via the adaptive MP to learn local information in the $G^{r,k}$, and (2) meanwhile those across mesh graphs to learn global information via up-sampling from coarse graphs $G^{r-1}$ to fine ones $G^{r}$.

{Finally, Alg. \ref{algorithm3} gives the Pseudocode of the overall processing steps of UA-MGN. The input includes the $R\times K$ pre-divided subgraphs (by Alg. \ref{algorithm1}) and the numbers of MP steps  (by Alg. \ref{algorithm2}). In lines 3-8, the \textbf{for} loop performs associated adaptive MP on each subgraph $G^{r,k}$. Note that the propagation of the $K$ subgraphs is independent and we thus perform parallel propagation for better speedup. Then, line 10-13 performs up-sampling for each non-final level. Finally, line 15 decodes the final node embedding vectors of the bottom graph $G^R$ (i.e., the input graph $G$ ) to generate the mechanical response output. 

Time complexity is $O(\frac{R\cdot K\cdot L^{r,k}}{\Theta})$ where $\Theta$ is the speedup 
ratio after parallel MP is adopted.

\begin{algorithm}[htbp]
    \caption{The Overall Processing Steps of UA-MGN}
    \label{algorithm3}
    \KwIn{$R\times K$ subgraphs $G^{1,1}, \ldots, G^{R,K}$, Steps $L^{1,1}, \ldots, L^{R,K}$ }
    \KwOut{Mechanical Response ${u\left ( \mathbf{v}_{i} \right ) }$ with $\mathbf{v}_{i} \in G^R$}

    Initiate $\mathbf{v}_{i}^{r}$ by node encoder for $1\leq r\leq R$;
    
    \For(\tcp*[f]{the $r$-th mesh graph}){$r=1$ to $R$} {

        \For(\tcp*[f]{the $k$-th subgraph}){$k=1$ to $K$} {
        
            Initiate $\vec{\mathbf{e}}_{ij}^{r,k,0}$ by edge encoder; $\mathbf{v}_{i}^{r,k,0} \leftarrow \mathbf{v}_{i}^{r}$;
        
            \For(\tcp*[f]{the $l$-th MP step}){$l=0$ to $L^{r,k}-1$} {
                Message propagation on $\mathbf{v}_{i}^{r,k,l}$ and $\vec{\mathbf{e}}_{ij}^{r,k,l}$; //\textbf{Eq. 5}
            }
        }
        
        Update $\mathbf{v}_i^r$ by aggregating $[\mathbf{v}^{r,1,L^{r,k}}_{i}, ..., \mathbf{v}^{r,K,L^{r,k}}_{i}]$; //\textbf{Eq. 6}

        \If{r < R} {
            Initiate $\vec{\mathbf{e}}_{ij}^{r,r+1}$ by edge encoder;
            
            Upd. $\mathbf{v}^{r+1}_{j}$ by up-sampling on $\vec{\mathbf{e}}^{r,r+1}_{ij}$ and $\mathbf{v}^{r}_{i}$; // \textbf{Eq. 4}
        }
    }

    
    \Return{$u\left ( \mathbf{v}_{i}\right ) \gets$ decode($\mathbf{v}^R_{i}$)}
\end{algorithm}

\section{Experiments}\label{sec5:experiment}
\subsection{Experimental Setup}
\subsubsection{Datasets}\label{sec511}
We use two synthetic and one real datasets.
\begin{itemize}
\item {\textbf{Beam}}. We generate the 2D Beam dataset by a widely used FEM solver ABAQUS\footnote{https://www.3ds.com/products/simulia} to simulate the deformation responses of rectangular beams subjected to an external force. The beams have the size of $15 \times 100$ $\mathrm{mm}^2$. Each beam contains a circular hole with a diameter of $5$ $\mathrm{mm}$. By varying the center of the hole, we generate 111 beam objects. That is, 
starting from the initial center $\left < 5,5 \right >\mathrm{mm}$ with the step size $2.5$ $\mathrm{mm}$, we move the center horizontally and vertically by 3 and 37 times, respectively. Meanwhile, we fix the bottom of the beam structure and then apply an external force {of} 300 $\mathrm{N}$ (Newton) to the top.  We also vary the force direction by changing the included angle between the force and the horizontal direction from $-60^{\circ}$ to $60^{\circ}$ with {a step of} $30^{\circ}$, generating 5 loading settings. For a given mesh graph, we solve the stress field on the associated Beam structure by the FEM solver ABAQUS as ground truth. 
\item {\textbf{SteeringWheel}}. The real dataset includes {239 samples of 3D steering wheels} provided by expert engineers from an automotive supplier. Following an industry trial standard, the engineers apply a force {of} 700 $\mathrm{N}$ in the negative $z$-axis direction {at the steering wheel rim} and meanwhile fix the {steering column} at the bottom plane. In each sample, we divide the steering wheel into hybrid mesh types of hexa-, penta- and tetra-hedral. The engineers then exploit an industry-level FEM solver LS-DYNA\footnote{https://lsdyna.ansys.com/} to generate numerical result of the stress field (as ground truth).

\item {\textbf{CylinderFlow}}. This synthetic dataset \cite{PFAFF2020} consists of time series of 2D mesh-based dynamic velocity evolution of {fluid} flow around a cylinder {as an obstacle}. Each mesh is with an associated time series. By varying the radius and centers of the obstacles, we have the associated simulation samples. {Since the dataset involves the multi-step time series data, we thus are interested in the \emph{1-step data} from $t=0$ to $t=1$, and the multi-step \emph{rollout data} from $t=0$ to $t=T$ for $T>1$.} 
\end{itemize}

Note that the Beam and CylinderFlow datasets are with the raw {geometries}, we can comfortably exploit the Delaunay triangulation \cite{10400424} to generate {coarse meshes} for $R$-level mesh graphs. Yet the {samples of the SteeringWheel dataset}  provided by the supplier are with one fine mesh graph without {the raw geometries}, we follow the work \cite{LINO2021} to generate coarse mesh graphs $G^1, ..., G^{R-1}$. 

For the Beam and SteeringWheel datasets, {we use} 80\% samples for training, 10\% samples for validation, and 10\% samples for testing. For the CylinderFlow dataset, we follow the settings in the previous work \cite{PFAFF2020} by using 1000 trajectories for training, 100 trajectories for validation, and 100 trajectories for testing. We perform baseline study on the three datasets. In terms of the remaining studies such as generalization and ablation study, we mainly use the Beam dataset, because we can comfortably change the initial and boundary conditions during the stress field simulation.

\newcolumntype{Y}{>{\centering\arraybackslash}X}
\begin{table}[htbp]
    \begin{tabularx}{0.5\textwidth}{cYYY}
        \toprule
        {Dataset} & {Beam} & {SteeringW.} & {CylinderF.}  \\
        \hline
       \textbf{\# of Samples} & 555 & 239 & 1200 \\
        \textbf{Mesh Type} & triangles & hexa-, penta-, tetra-hedra & triangles \\
        \textbf{Nodes per Sample} & 522.77 & 72061.01 & 1885.06 \\
        \textbf{Edges per Sample} & 1444.32 & 200525.86 & 5420.65 \\
        \textbf{Max/Min/Avg. Degree} & 8/3/5.53 & 39/3/5.55 & 8/2/5.57 \\
        \bottomrule
    \end{tabularx}
    \caption{Overview of Three Datasets}\vspace{-4ex}
    \label{tab:datasets}
\end{table}

\subsubsection{Baselines}\label{sec5.1.2baseline}
\begin{itemize}
\item {\textbf{UNet \cite{NILS202025}}}: We use a U-shaped CNN model to learn multi-scale physical system simulation.

\item {\textbf{FNO \cite{LI2020}}}: A Neural Operator approach using Fourier space to learn complex patterns and correlations.

{
\item {\textbf{Geo-FNO \cite{10.5555/3648699.3649087}}}: A very recent geometry-aware improved version of FNO to learn deformation from the irregular input mesh to a latent uniform grid to avoid the limitations of {Fourier transforms}.
}

\item {\textbf{MGN \cite{PFAFF2020}}}: A flat GNN model to learn mesh-based simulations. The GNN is to represent the spatial relationships within mesh graphs.

\item {\textbf{MS-MGN \cite{FORTUNATO2022}}}: The state-of-the-art  work essentially is a hierarchical version of MGN with multi-stacked U-shapes to represent fine and coarse meshes. 

\item {\textbf{AMR-GNN \cite{PERERA2024117152}}}: A very recent U-shaped GNN model with multigraphs consist of uniform grids of various resolutions to learn features at different scales.

\end{itemize}

Note that UNet and FNO require very regular {input} structures. Yet, for our mesh graph data, particularly for the SteeringWheel samples with rather complex structure, these two approaches cannot work well, and we do not have the evaluation results of UNet and FNO on the SteeringWheel data. For fairness, we use the \emph{equal number of total MP steps} for MGN, ours and MS-MGN (including the MP steps in MGN, up-sampling steps in ours, and the up- and down-sampling steps in MS-MGN).

{\subsubsection{Evaluation Metric} We measure the performance by Root Mean Square Error with $\mathrm{RMSE}=\sqrt{\frac{1}{N} {\textstyle \sum_{i=1}^{N} \frac{1}{n_{i}} \textstyle \sum_{j=1}^{n_{i}}(y_{ij}-\hat{y}_{ij} )^2}}$, where $N$ is the number of testing samples and $n_i$ is the number of nodes in $i$-th sample, and $y_{ij}$ is the ground truth value while $\hat{y}_{ij}$ is the prediction value of the $j$-th node in the $i$-th sample.}

We implement our work UA-MGN by Python 3.10 and PyTorch 2.1.0, and all experiments are performed on a server with a 16-core Xeon(R) Gold 6430 CPU and an NVIDIA GeForce RTX 4090 GPU.



\vspace{-1ex}

\subsection{Performance Study}

\subsubsection{Baseline Study}
Figure \ref{fig:overall} first gives the baseline study. Among all approaches, our work UA-MGN performs best on three datasets with the following findings. 
\begin{itemize}
    \item In the Beam dataset, the four hierarchical GNN models (UNet, AMR-GNN, MS-MGN and ours) and two neural operator models (FNO and Geo-FNO)
can learn wide or global receptive fields and lead to better performance than the single-layer flat GNN model (MGN). Similar results occur in the CylinderFlow 1-step and Rollout scenarios.

\begin{figure}[htbp]
  \centering
  \begin{subfigure}{0.45\linewidth}
    \includegraphics[width=.9\linewidth]{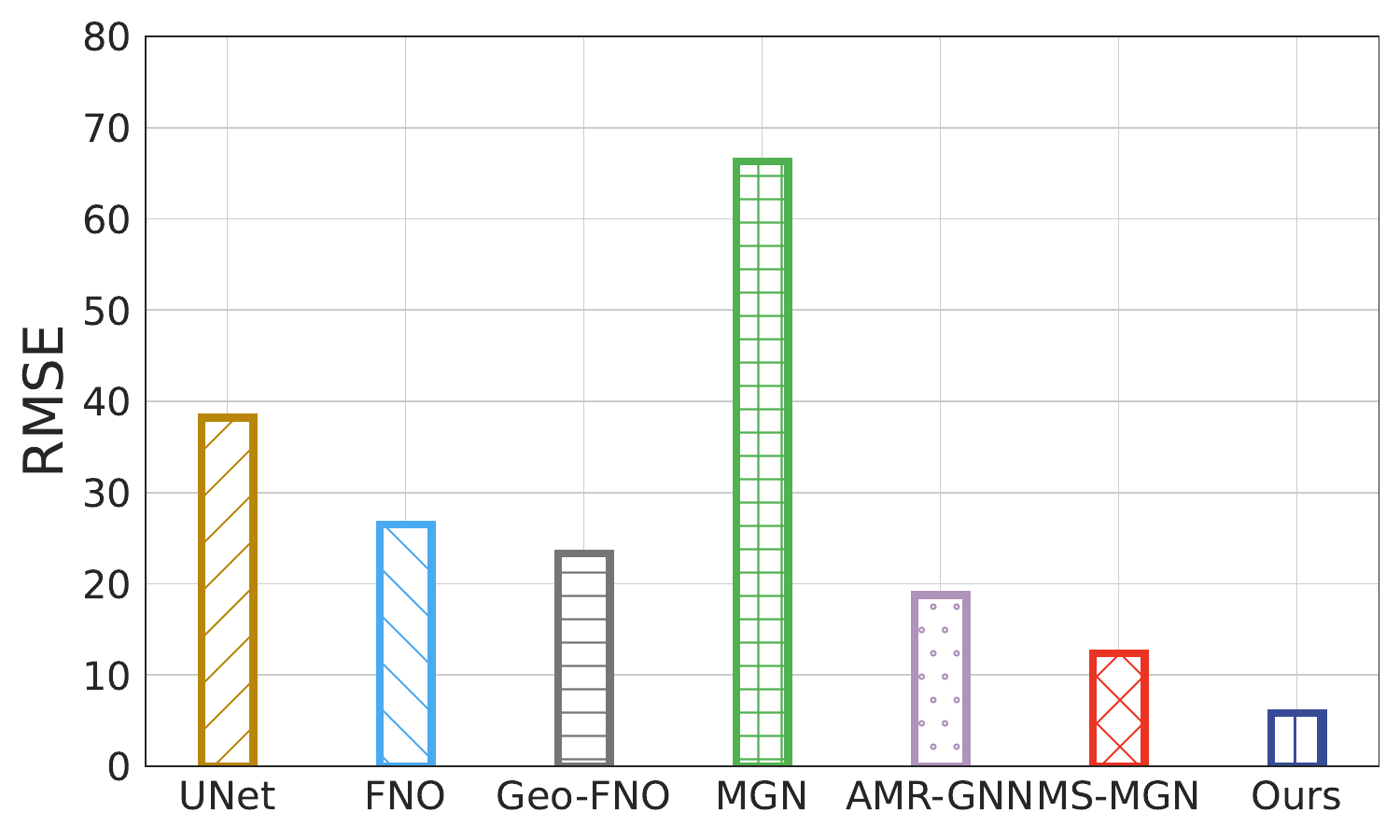}\vspace{-1ex}
    \caption{Beam}
    \Description{}
  \end{subfigure}
  \begin{subfigure}{0.45\linewidth}
    \includegraphics[width=.9\linewidth]{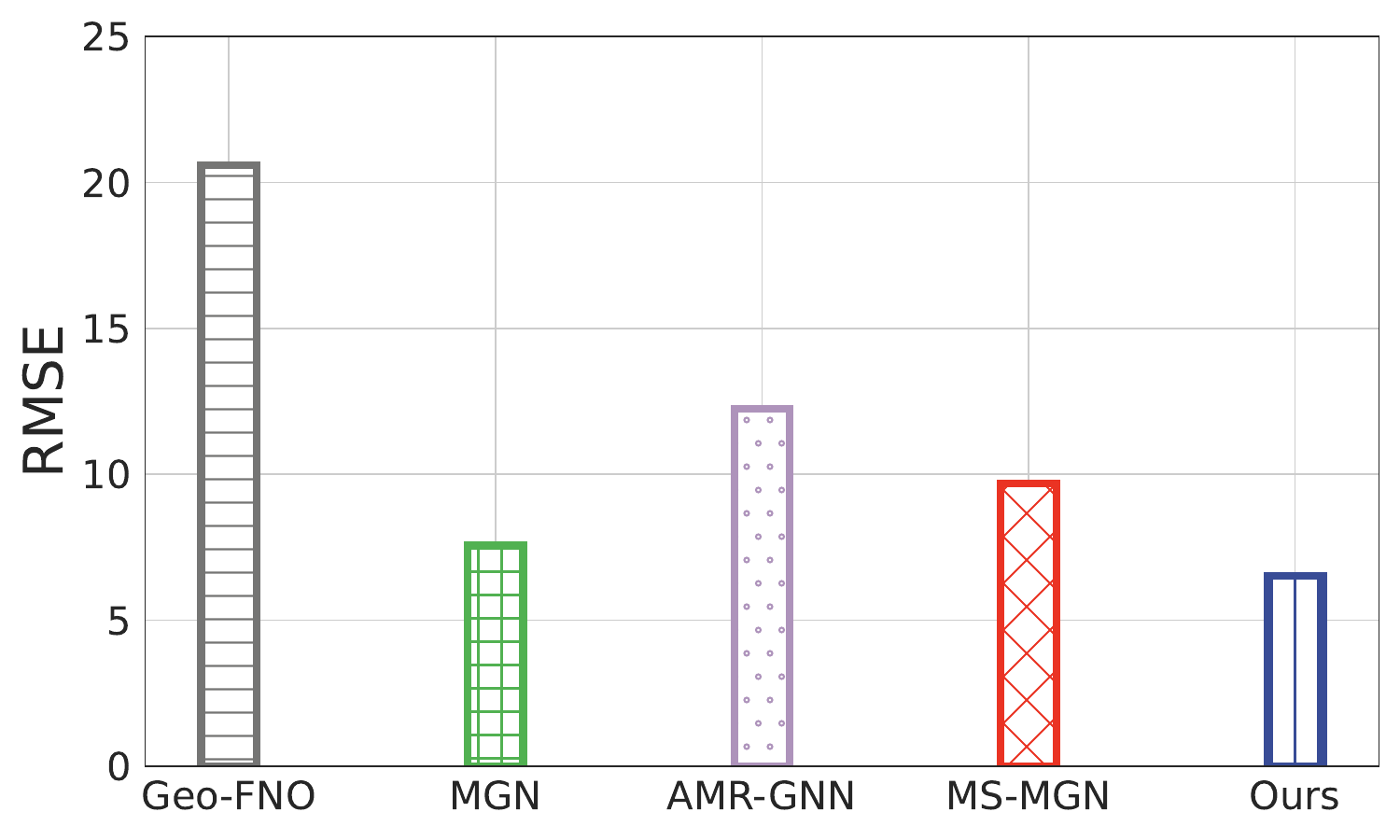}\vspace{-1ex}
    \caption{SteeringWheel}
    \Description{}
  \end{subfigure}
  
  \begin{subfigure}{0.45\linewidth}
    \includegraphics[width=.9\linewidth]{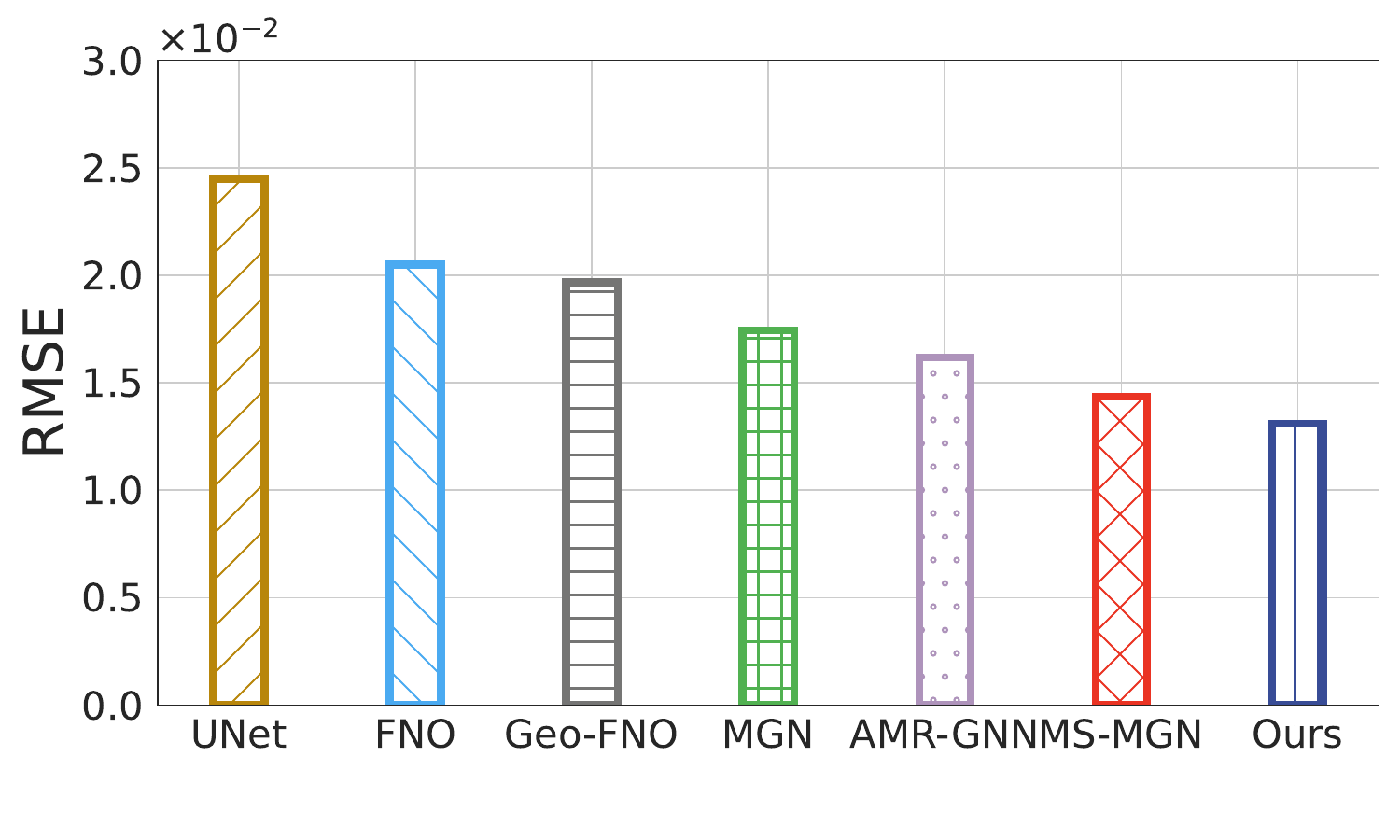}\vspace{-1ex}
    \caption{CylinderFlow 1-step}
    \label{subfig:cylinderflow 1-step}
    \Description{}
  \end{subfigure}
  \begin{subfigure}{0.45\linewidth}
    \includegraphics[width=.9\linewidth]{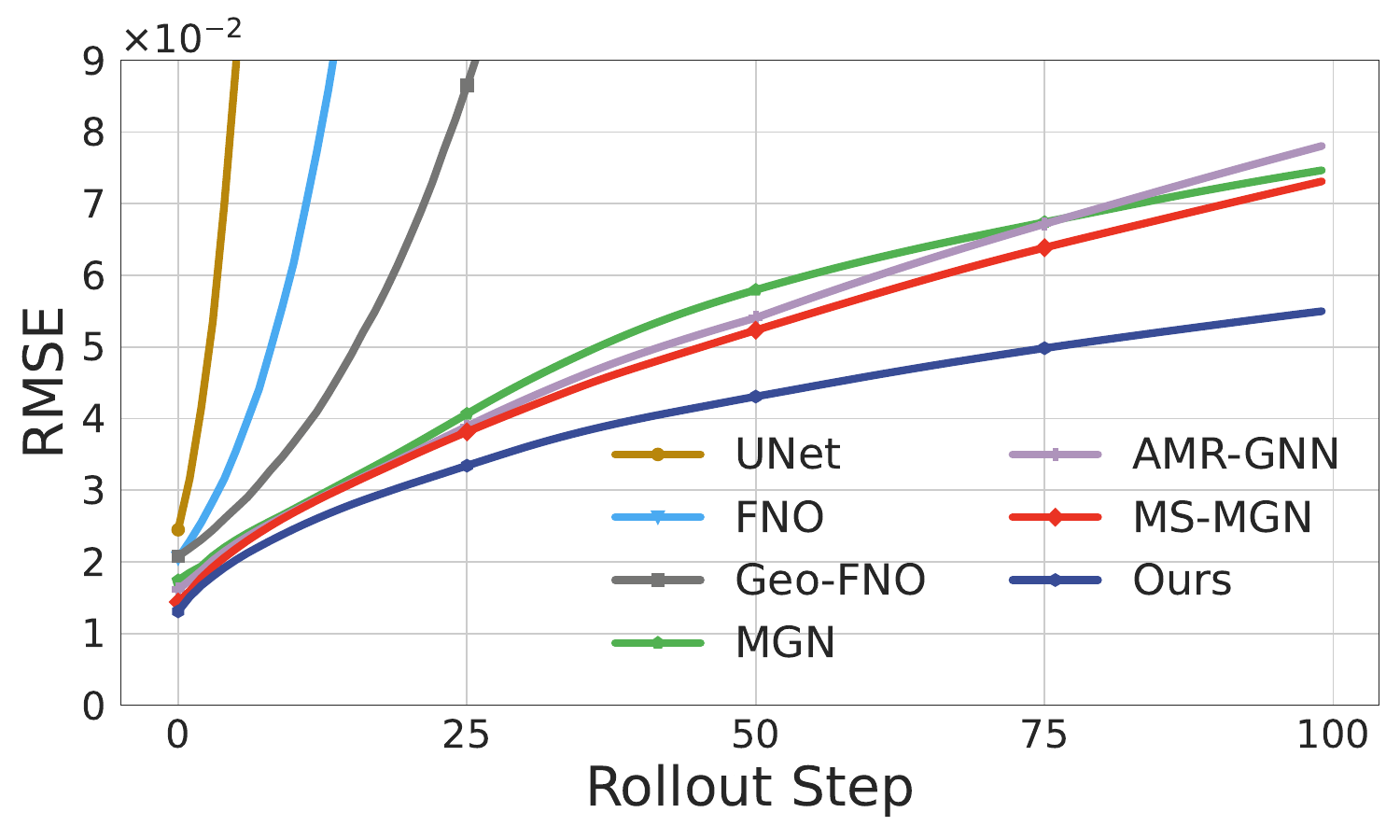}\vspace{-1ex}
    \caption{CylinderFlow Rollout}
    \Description{}
  \end{subfigure}\vspace{-2ex}
  \caption{Baseline Study}\vspace{-1ex}
  \label{fig:overall}
     \vspace{-1ex}
\end{figure}

\item Note that UNet and FNO do not work on the {SteeringWheel} dataset due to complex geometric structures (see Section \ref{sec5.1.2baseline}). We thus plot the results of the five remaining approaches. In this dataset, Geo-FNO, AMR-GNN and MS-MGN do not achieve better results than MGN, mainly due to the complex geometric structures of steering wheels, and it is hard for Geo-FNO to exploit a learning model for the transformation from the input  mesh structure to the uniform grid. The three remaining works, AMR-GNN, MS-MGN and ours, all require the coarsen grids generated by the work \cite{LINO2021}, which unfortunately cannot precisely represent global geometric boundaries. Instead, our work UA-MGN interweaves the multi-level node feature representation (via MP) and feature concatenation-aggregation for better global and local feature representation.

 \item Finally, in the two CylinderFlow settings, we note that this CylinderFlow simulation involves rather complex interactions between fluid flow and the cylinder. UNet, FNO and Geo-FNO do not work well {in learning} such complex interactions, mainly because the three approaches represent multi-scale features independently before feature aggregation. Instead, the four GNN-based models (MGN, AMR-GNN, MS-MGN and ours) leverage mesh graphs to better learn complex interactions, and particularly our work UA-MGN performs best with the lowest RMSE.
\end{itemize}

\subsubsection{Generalization Study}
In this section, we vary the geometric structure or boundary conditions of the Beam objects to evaluate the generalization ability of our work against two mesh graph models (MGN and MS-MGN). Here, we still use the training data from Section \ref{sec511}, and yet generate the alternative testing data by changing the  \emph{geometric structure} including (1) the shapes of the hole to squares or regular hexagons with the same diameter, (2) the positions of the hole center, starting from the initial position $\left <6.25,6.25 \right >\mathrm{mm}$ {with a step of} $2.5 \mathrm{mm}$ to move the hole center in the horizontal and vertical directions with $2\times 36$ steps, (3) the number of the holes by randomly choosing two non-overlapping holes, and (4) the diameters of the hole to 4 or 6 $\mathrm{mm}$. In addition, we change the \emph{boundary conditions} by (1) the force directions ranging from $-75^{\circ}$ to $75^{\circ}$ with the step $30^{\circ}$, (2) the force values to either 270 or 330 $\mathrm{N}$, and (3) the force position to $y=90\mathrm{mm}$, respectively.

In Figures \ref{fig:generalstudy1} and \ref{fig:generalstudy2}, our work UA-MGN consistently outperforms the two competitors when given various geometric structures and boundary conditions, indicating better generalization ability.

\begin{figure}[t]
    \centering
  \begin{subfigure}{0.45\linewidth}
    \includegraphics[width=.9\linewidth]{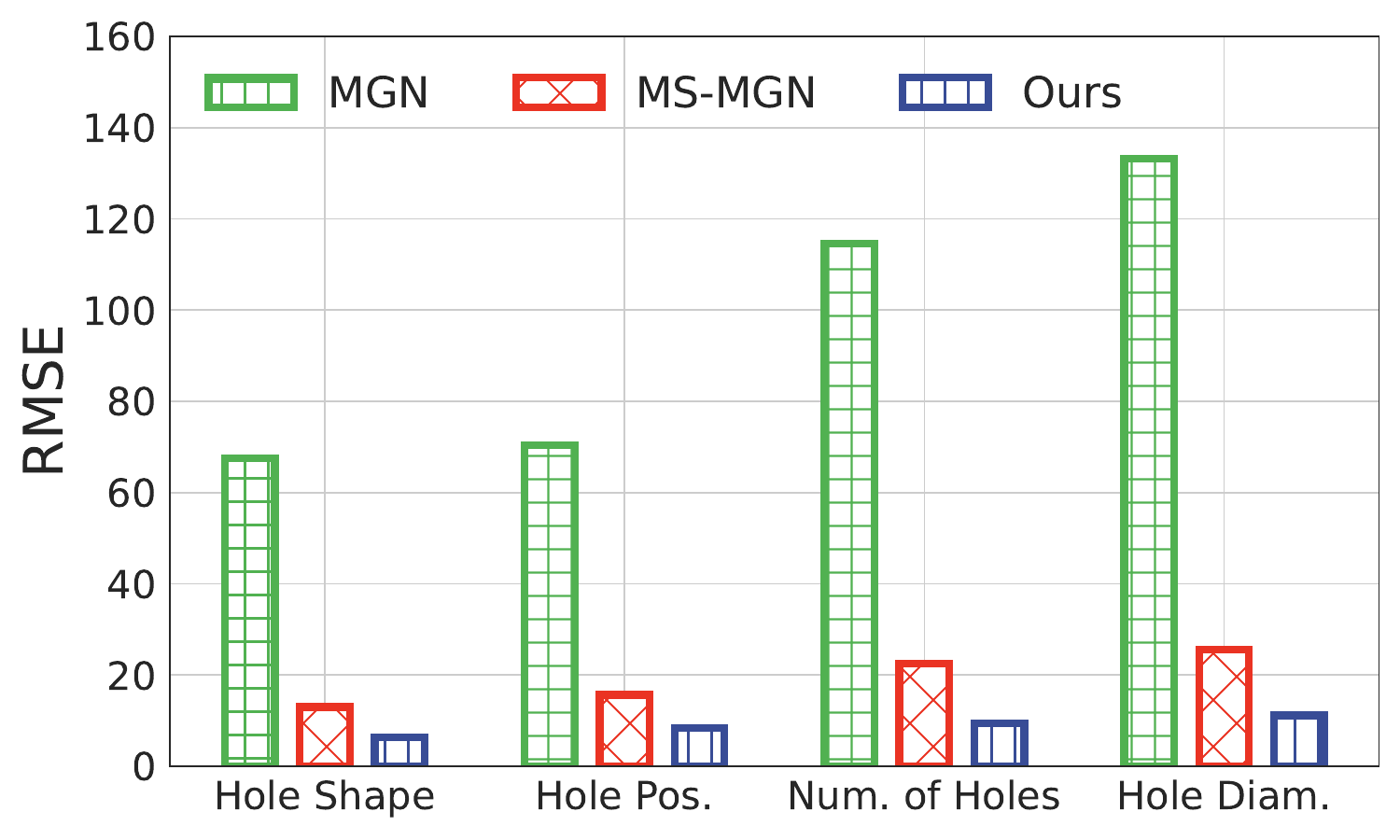}\vspace{-1ex}
    \caption{Geometric Structure}
        \label{fig:generalstudy1}
  \end{subfigure}
  \begin{subfigure}{0.45\linewidth}
    \includegraphics[width=.9\linewidth]{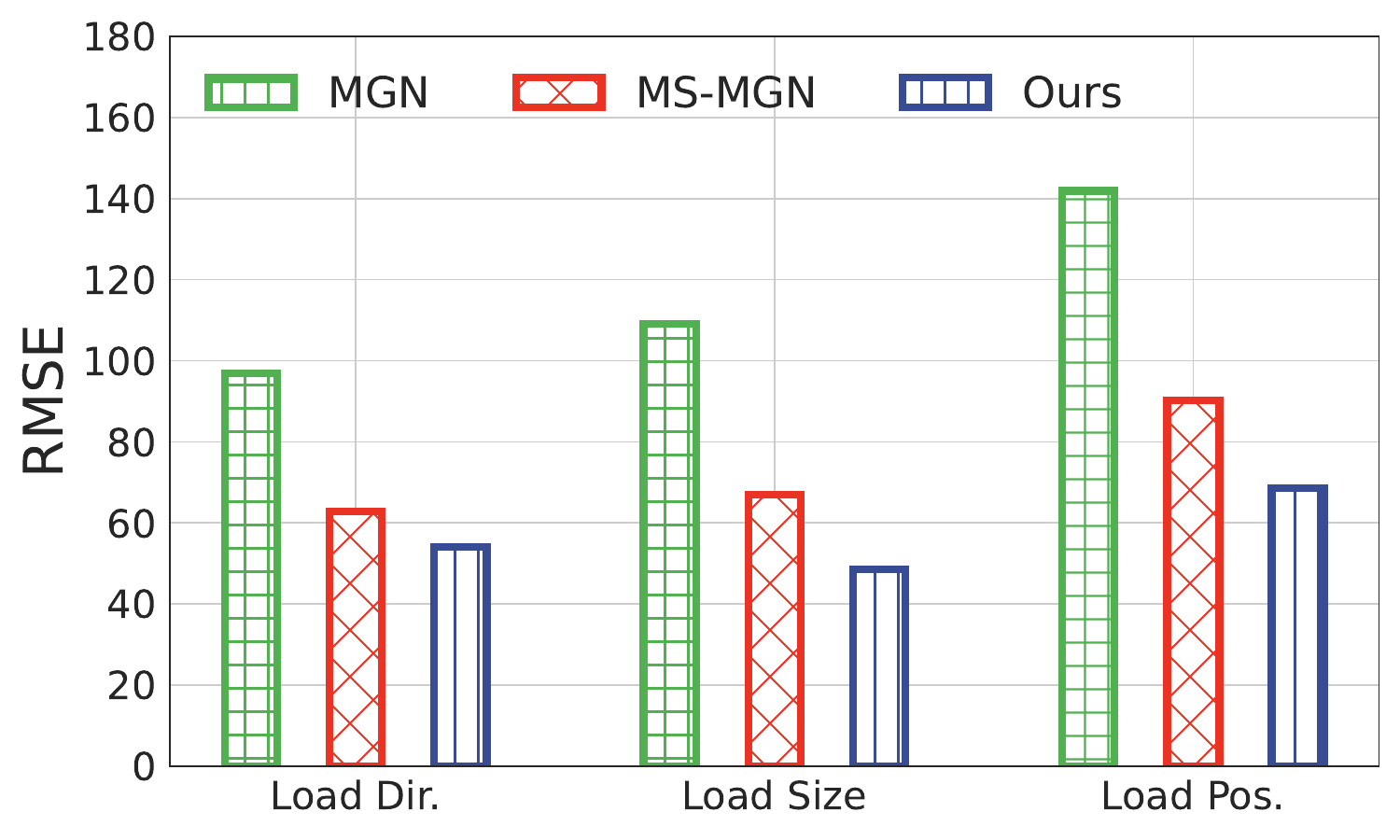}\vspace{-1ex}
    \caption{Boundary Condition}
    \label{fig:generalstudy2}
  \end{subfigure}
\\    
    \begin{subfigure}[t]{0.45\linewidth}
        \includegraphics[width=.9\linewidth]{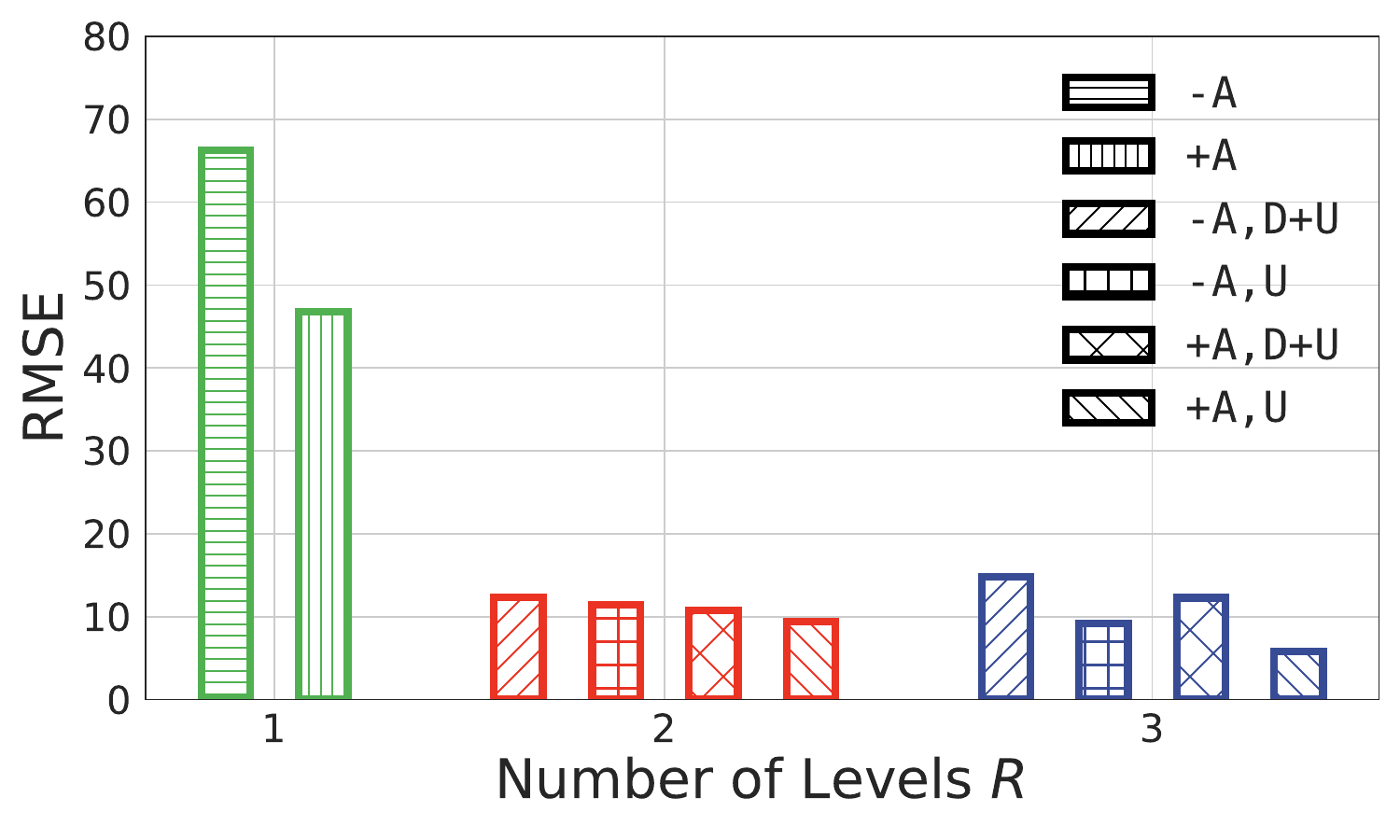}\vspace{-1ex}
        \caption{Ablation Study}
        \label{fig:ablation}
    \end{subfigure}
    \begin{subfigure}[t]{0.45\linewidth}
        \includegraphics[width=.9\linewidth]{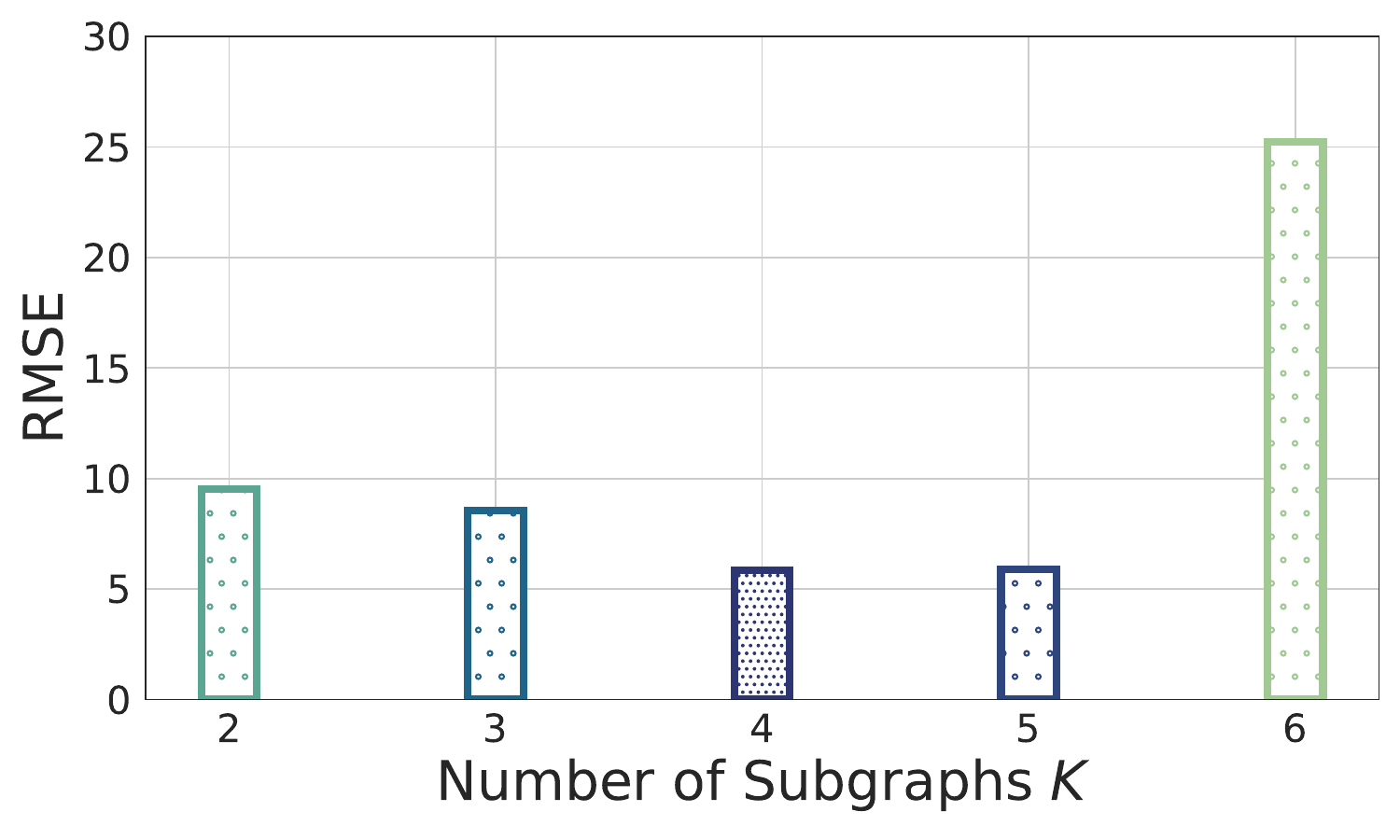}\vspace{-1ex}
        \caption{Effect of Subgraphs}
        \label{fig:direction}
    \end{subfigure}\vspace{-2ex}
    \caption{(a-b) Generalization Study, (c) Ablation Study, and (d) Sensitivity Study}     \vspace{-1ex}\label{fig:ablation5}    \vspace{-1ex}   
\end{figure}

\subsubsection{Ablation Study}
To study the benefits of two proposed techniques: the up-sampling-only mesh graph networks and adaptive message propagation (MP), we have four variants in Figure \ref{fig:ablation}. For example, $\langle  \mathsf{+A, U+D} \rangle$ indicates the UA-MGN variant with \underline{a}daptive MP and the \underline{u}p- and \underline{d}own-sampling, $\langle\mathsf{-A, U}\rangle$ {is} the variant without the adaptive MP and yet with the up-sampling only, and $\langle\mathsf{+A, U}\rangle$ is just our UA-MGN model with the adaptive MP and the up-sampling only. In addition, we are interested in how these variants perform on the $R$-level mesh graphs with various levels $R$. From Figure \ref{fig:ablation}, we have the following result.

\begin{itemize}
\item In terms of adaptive MP, the variants without adaptive propagation consistently lead to higher RMSE. {It is mainly because the adaptive MP makes the models more adaptable to various geometric structures}.
\item The variants with down- and up-sampling steps suffer from much higher RMSE than those with up-sampling only. That is, besides better global representation at an early stage, the variants with up-sampling only (including ours) allow more MP steps within each mesh graph to better learn local features than the {variants} with down- and up-sampling. 
\item When the number $R$ grows from 1 to 3, the RMSE of all UA-MGN variants becomes smaller, indicating that multi-level mesh graphs do help better representation of the geometric structures. 
Intuitively, MGN can be treated as the {variant with $R=1$ and without the adaptive MP}.




 \end{itemize}

\vspace{-1ex}

\subsection{Sensitivity Study}
In this section, we investigate the effect of some key parameters including the number of divided subgraphs $K$, the number $L^{r,k}$ of {MP} steps, and the total {number of} MP steps.

\subsubsection{Number of Divided Subgraphs $K$}
We study the effect of divided subgraphs $K$ in Figure \ref{fig:ablation5}d. When the value of $K$ ranges from 2 to 6, the number $K=4$ leads to the smallest RMSE. It makes sense: since our algorithm performs the mesh edge-based division, too small a number, i.e., $K=2$, may not effectively differentiate edge directions. It is particularly true that the Beam data samples consist of triangle meshes. Instead, too high a number, i.e., $K=6$, could reversely aggravate the benefit of adaptive MP: a greater number $K=6$ indicates more diverse MP directions and too small subgraphs, such that the MP is limited within such very small 
subgraphs and harms the sufficient range of receptive field, leading to worse representation of global features. Consequently, the number $K=4$ can best balance the MP directions and sufficient range of receptive field, and leads to the smallest errors.

\begin{figure}[h]
  \begin{subfigure}{0.45\linewidth}
    \includegraphics[width=1.0\linewidth]{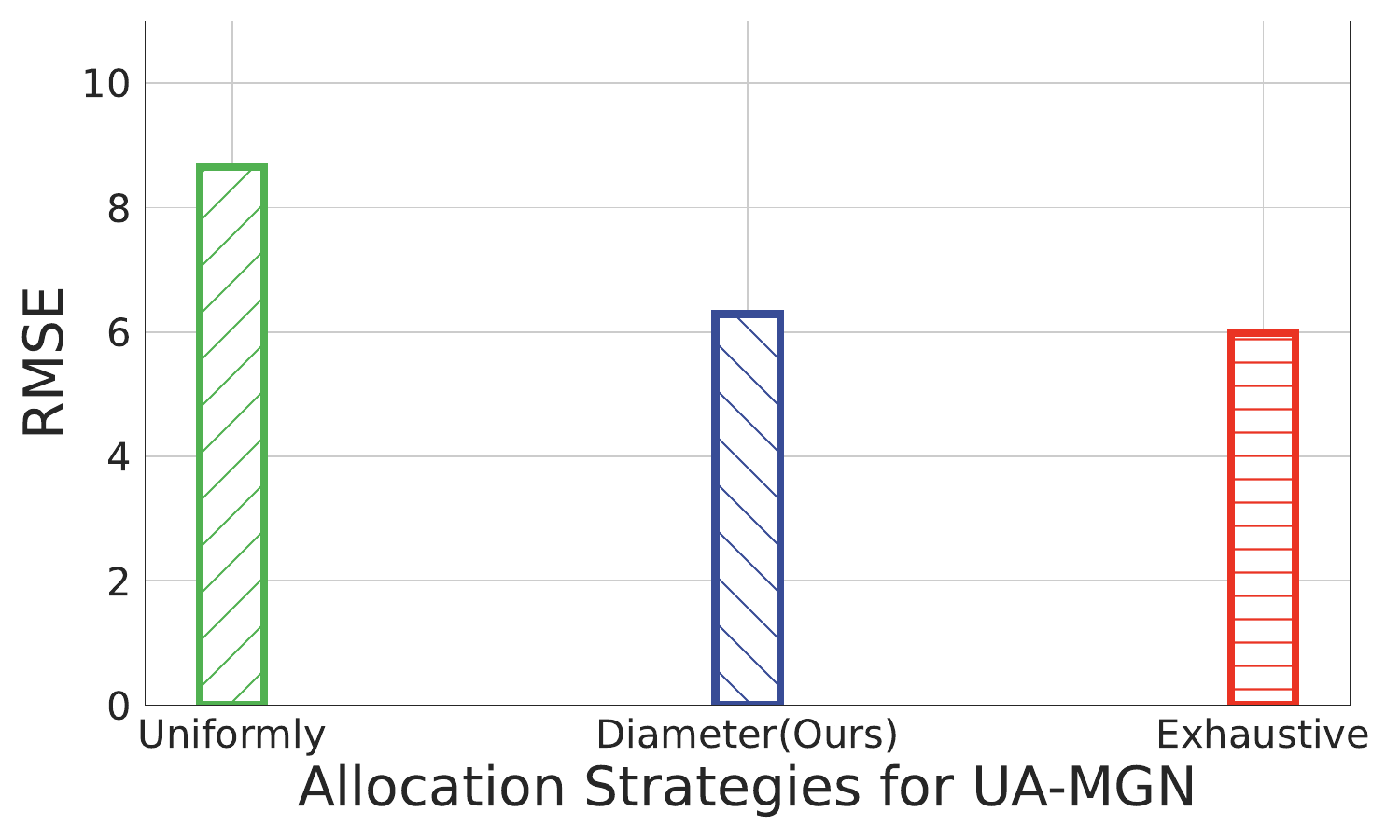}\vspace{-1ex}
    \caption{Setting MP Steps $L^{r,k}$}
    \label{subfig:strategies}
    \Description{}
  \end{subfigure}
  \begin{subfigure}{0.45\linewidth}
    \includegraphics[width=1.0\linewidth]{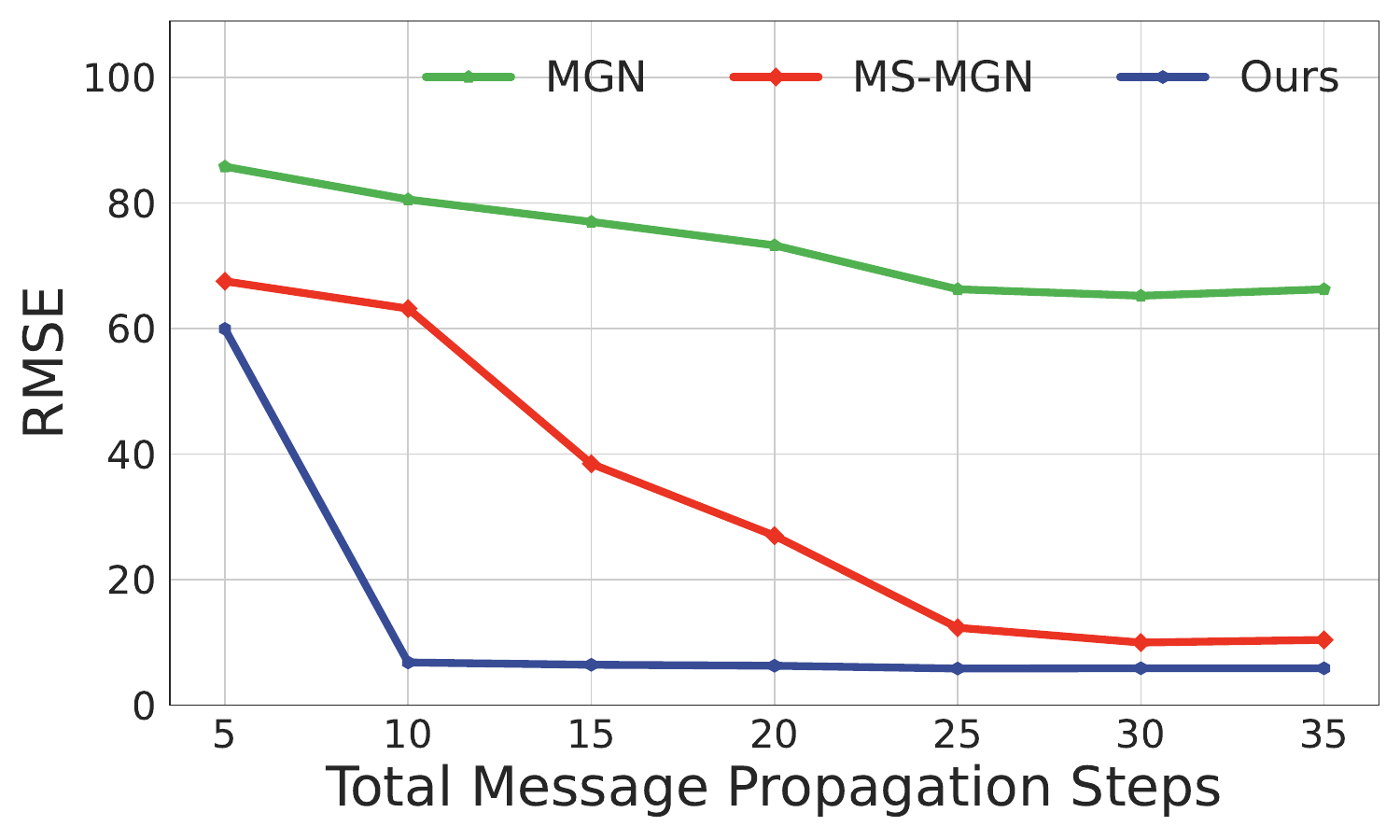}\vspace{-1ex}
    \caption{Total MP Steps}
    \label{subfig:total}
    \Description{}
  \end{subfigure}
  \\
  \begin{subfigure}{0.45\linewidth}
    \includegraphics[width=1.0\linewidth]{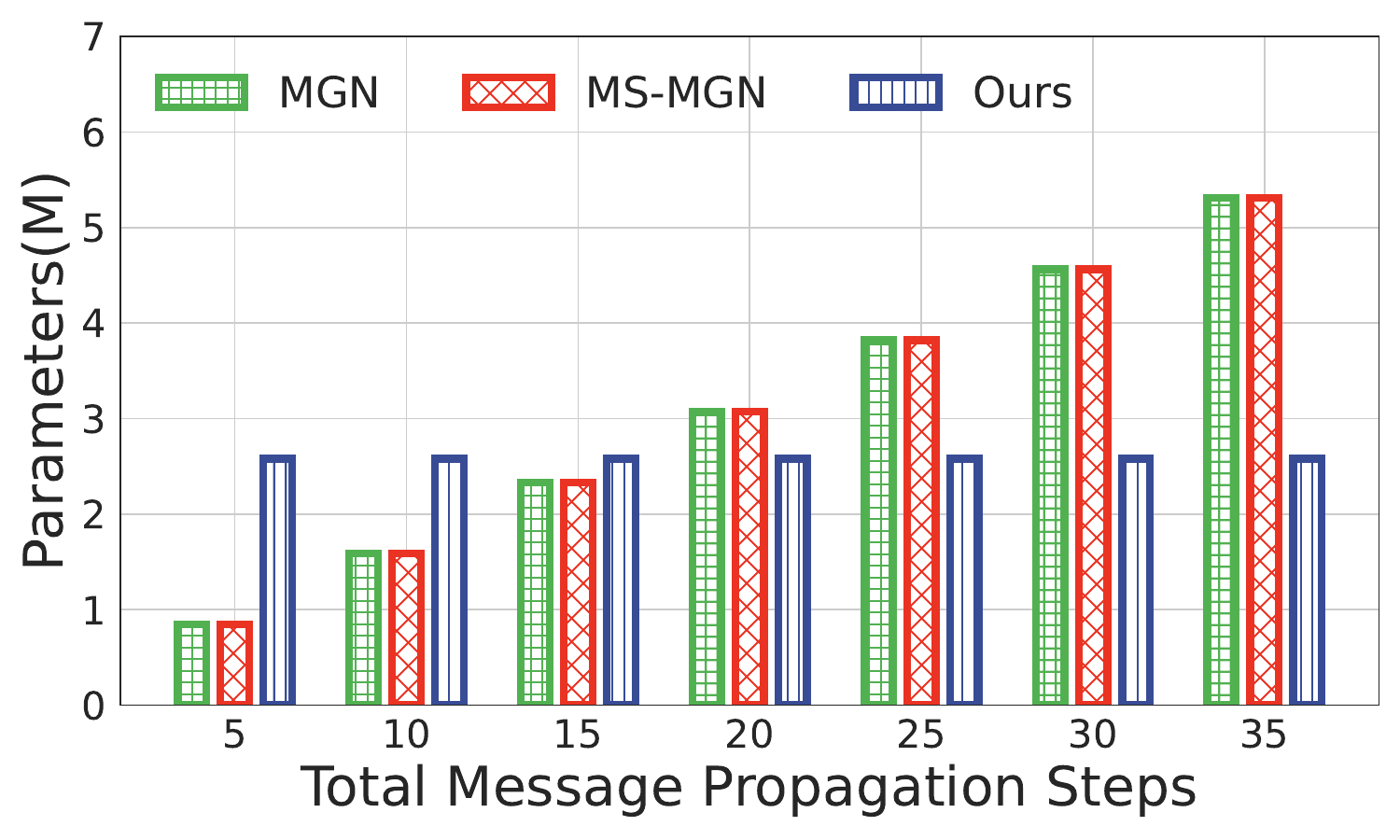}\vspace{-1ex}
    \caption{Network Parameters}
    \label{subfig:complexity}
    \Description{}
  \end{subfigure}
  \begin{subfigure}{0.45\linewidth}
    \includegraphics[width=1.0\linewidth]{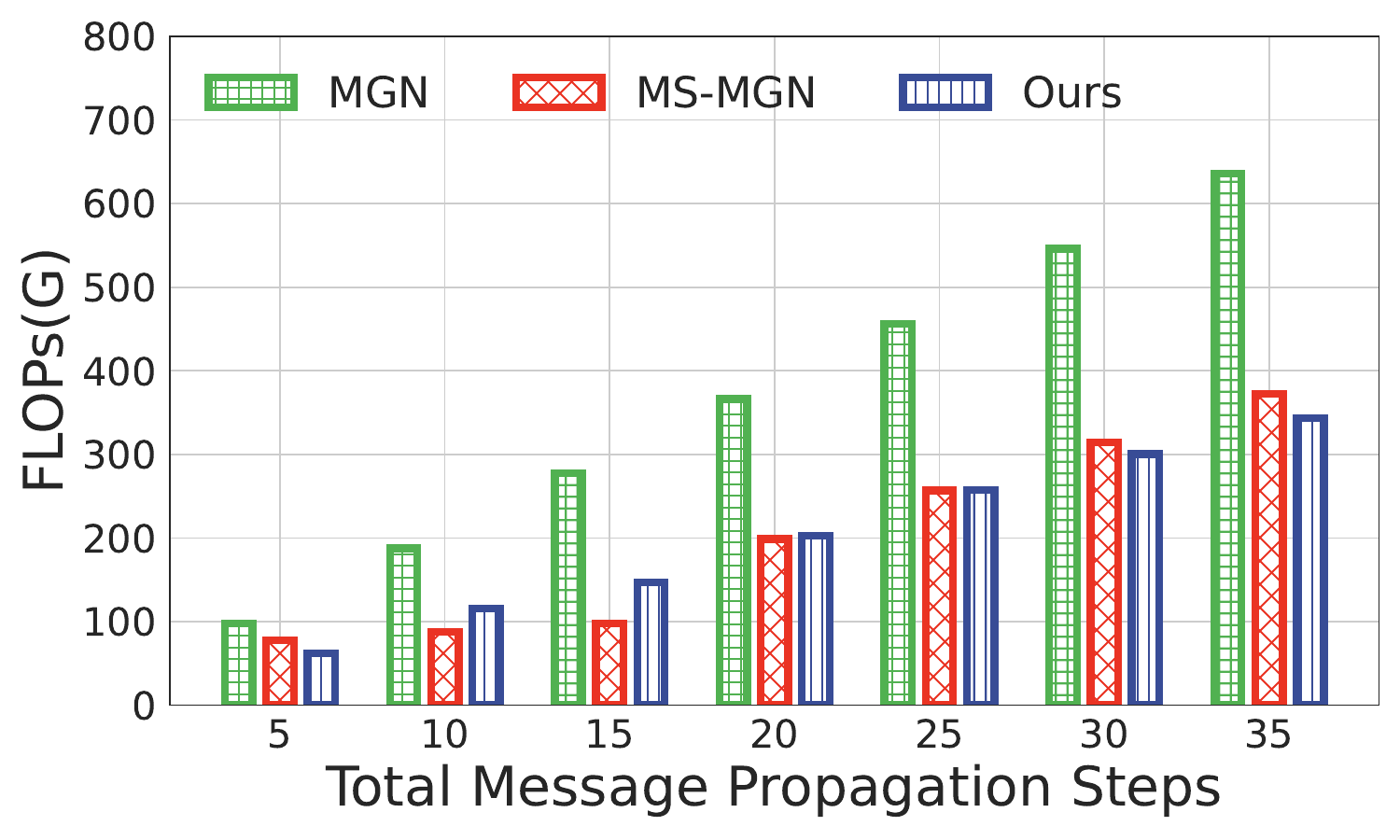}\vspace{-1ex}
    \caption{Computation Cost}
    \label{subfig:efficiency}
    \Description{}
  \end{subfigure}\vspace{-2ex}
  \caption{ (a, b) Sensitivity Study,  (c, d) Efficiency Study}
  \label{fig:MP Steps}
\end{figure}

\subsubsection{Tuning the number $L^{r,k}$ of MP steps }
Recall that we use the maximal diameter of some connected {components (CCs)} to set the number $L^{r,k}$. We consider the alternative approaches to tune $L^{r,k}$.
\begin{itemize}
    \item \textbf{Uniform}: We assign  
    an equal {number of} MP steps for all subgraphs within every level mesh graph. 
    \item \textbf{Diameter (ours)}: Section \ref{sec4.2} uses the maximal diameter of {CCs within} those sub-areas to tune $L^{r,k}$.
    \item \textbf{Exhaustive}: The exhaustive search is to find the best $L^{r,k}$, and we compare how ours is close to this best one. 
\end{itemize}

In Figure \ref{subfig:strategies}, the uniform approach simply assigns the same {number of} MP steps, and may not consider the effects of MP directions and mesh graph levels, suffering from the highest errors. Instead, the errors of ours are rather close to the exhaustive result. Nevertheless, the exhaustive approach is {time-consuming due to the} exponential search space, {with} the training time {for} an individual instance of UA-MGN with the specific $L^{r,k}$ numbers is around 1 hour. Yet, our approach requires only 23.76 seconds. 



\subsubsection{Total Number of MP steps }\label{se5.3.3}
We study the effect of total {number of} MP steps in our work against two mesh graph-based competitors (MGN and MS-MGN). For our work, this number involves the MP within {mesh graph at each level} and up-sampling across mesh graphs. Yet, MS-MGN takes into account the MP within {mesh graph at each level}, up-sampling and down-sampling across mesh graphs. Since MGN does not involve {up- and down-} sampling operations, we only have MP within a single mesh graph. Figure \ref{subfig:total} indicates that, overall, a higher total number of MP steps leads to lower RMSE and finally converges to a stable value after the MP steps become larger than 30. Such results clearly demonstrate that our work UA-MGN can converge the fastest to the lowest error at an early stage with only 10 total MP steps.


\subsection{Efficiency Study}
In this experiment, we measure network parameters and computation cost of UA-MGN against MGN and MS-MGN. Note that for a certain number of MP steps, MGN and MS-MGN require an associated MLP to perform the MP namely one MP block. Thus, by varying the total MP steps from 5 to 35, MGN and MS-MGN need to build the 5-35 MP blocks (i.e., MLPs), requiring the corresponding network parameters for these MLPs.

In Figure \ref{subfig:complexity}, when the number of MP steps varies from 5 to 35, the network parameters of MGN and MS-MGN grow from nearly 1 million to 5.3 million. Yet the network parameters of UA-MGN are fixed with no change, with the parameters of only 2.6 million. It is mainly because (1) we \emph{share} the MLP blocks for {the same divided subgraph}, independent {of} the MP steps, and (2) the number $K=4$ of divided subgraphs is much smaller than the number of MP steps, such as 35. 
From Figures \ref{subfig:total} and \ref{subfig:complexity}, 
the key insight of this experiment is that when the errors of all three approaches become stable (e.g., 30 MP steps), the two competitors suffer from much higher errors though with more parameters than ours. For example, for 30 MP steps, UA-MGN leads to 40.99\% lower errors meanwhile using only 43.48\% fewer network parameters than MS-MGN.}

Figure \ref{subfig:efficiency} plots the floating point operations (FLOPs) \cite{HOCKNEY1988} of three approaches. Using higher MP steps from 5 to 35 require more computation cost, i.e., higher FLOPs. Here, our work and MS-MGN require much lower FLOPs than MGN. For 30 MP steps, UA-MGN uses 4.49\% fewer FLOPs than MS-MGN. The behind rationale is as follows. (1) Representation of coarse meshes requires lower FLOPs than fine meshes, and (2) all MP steps of MGN are on fine meshes, yet our work and MS-MGN involve coarse and fine meshes, thus leading to much smaller FLOPs. 

{Besides}, we are interested in the RMSE and the used FLOPs of three approaches when given a certain number of MP steps. From Figures \ref{subfig:total} and \ref{subfig:efficiency}, for the MP steps, e.g., 10, we find that UA-MGN can achieve the fastest convergence with the least FLOPs and smaller RMSE. Such result demonstrates the superiority of our work with higher prediction accuracy and lower computation cost.

Finally, regarding running time (measured by an average per sample) of the two datasets of {Beam} and {SteeringWheel} on our server, the FEM solver requires {an} average solving time {of} 600 $\mathrm{ms}$ and 20 $\mathrm{min}$, and UA-MGN only takes the prediction time {of} 1.23 $\mathrm{ms}$ and 201.8 $\mathrm{ms}$, respectively, faster than {that} of MGN (2.10 $\mathrm{ms}$ and 247.42 $\mathrm{ms}$) and MS-MGN (1.75 $\mathrm{ms}$ and 221.14 $\mathrm{ms}$). For training time, UA-MGN uses 6.42 $\mathrm{s}$ and 1.01 $\mathrm{min}$, respectively, still faster than MGN (10.63 $\mathrm{s}$ and 2.47 $\mathrm{min}$) and MS-MGN (6.92 $\mathrm{s}$ and 1.15 $\mathrm{min}$). Note that the FEM solver re-computes the entire simulation, whenever the samples change in terms of the geometric structure or the external force. Yet, learning-based models lead to significantly faster prediction time with only one training run.

\vspace{-1ex}

\section{Conclusion and Future Work}\label{sec6:conclude}
In this paper, we propose a novel hierarchical mesh graph network UA-MGN. The up-sampling only technique across multi-level mesh graphs leads to better global receptive fields and much smaller {MP} steps. The adaptive MP within a mesh graph can allow the MP along edge groups to overcome the issue of infinite MP loops and over-smoothing. Extensive evaluation on two synthetic and one real datasets demonstrates that UA-MGN outperforms state-of-the-art MS-MGN \cite{FORTUNATO2022} and the very recent works Geo-FNO \cite{10.5555/3648699.3649087} and AMR-GNN \cite{PERERA2024117152} with smaller prediction errors and higher efficiency. 
As future work, we are interested (1) the learning model of mesh generation for complex mechanical systems and (2) distributed simulation model for very large mesh graphs with tens of millions and even more nodes.

\newpage


\end{document}